\documentclass{article} 
\usepackage{iclr_conference,times}
\usepackage{amsmath,amsfonts,bm}

\usepackage{url}

\usepackage{graphicx}
\usepackage{xspace}
\usepackage{wrapfig}
\usepackage{multirow}
\usepackage{array}
\usepackage{booktabs}
\usepackage{tabularray}
\usepackage{algorithm}
\usepackage{algorithmic}
\usepackage{subcaption}
\usepackage[bookmarks=false]{hyperref}
\usepackage[capitalise]{cleveref}

\DeclareMathOperator*{\argmax}{arg\,max}

\title{Rethinking Visual Counterfactual \\ Explanations Through Region Constraint}

\author{Bartlomiej Sobieski \thanks{Corresponding author} \\
University of Warsaw\\
\texttt{b.sobieski@uw.edu.pl} \\
\And
Jakub Grzywaczewski \\
Warsaw University of Technology\\
\texttt{jakub.grzywaczewski2.stud@pw.edu.pl} \\
\And
Bartlomiej Sadlej \\
University of Warsaw\\
\texttt{b.sadlej@student.uw.edu.pl} \\
\And
Matthew Tivnan \\
Harvard Medical School\\
\texttt{mtivnan@mgh.harvard.edu} \\
\AND
Przemyslaw Biecek \\
University of Warsaw, Warsaw University of Technology \\
\texttt{przemyslaw.biecek@gmail.com}
}

\makeatletter
\DeclareRobustCommand\onedot{\futurelet\@let@token\@onedot}
\def\@onedot{\ifx\@let@token.\else.\null\fi\xspace}

\iclrfinalcopy 
\begin{document}

\maketitle

\begin{abstract}
Visual counterfactual explanations (VCEs) have recently gained immense popularity as a tool for clarifying the decision-making process of image classifiers. This trend is largely motivated by what these explanations promise to deliver -- indicate semantically meaningful factors that change the classifier's decision. However, we argue that current state-of-the-art approaches lack a crucial component -- the \emph{region constraint} -- whose absence prevents from drawing explicit conclusions, and may even lead to faulty reasoning due to phenomenons like confirmation bias. To address the issue of previous methods, which modify images in a very entangled and widely dispersed manner, we propose \emph{region-constrained} VCEs (RVCEs), which assume that only a predefined image region can be modified to influence the model's prediction. To effectively sample from this subclass of VCEs, we propose \emph{Region-Constrained Counterfactual Schrödinger Bridges} (RCSB), an adaptation of a tractable subclass of Schrödinger Bridges to the problem of conditional inpainting, where the conditioning signal originates from the classifier of interest. In addition to setting a new state-of-the-art by a large margin, we extend \method{} to allow for \emph{exact} counterfactual reasoning, where the predefined region contains \emph{only} the factor of interest, and incorporating the user to actively interact with the RVCE by predefining the regions manually.
\end{abstract}

\section{Introduction}

Visual counterfactual explanations (VCEs) aim at explaining the decision-making process of an image classifier by modifying the input image in a semantically meaningful and minimal way so that its decision changes. Over time, they have become an independent research direction with the latest methods presenting impressive and visually appealing results. Nevertheless, in this work we show that they possess a fundamental flaw at a conceptual level -- the lack of \emph{region constraint} and its proper utilization.

\begin{wrapfigure}{r}{0.5\linewidth}
    \includegraphics[width=\linewidth]{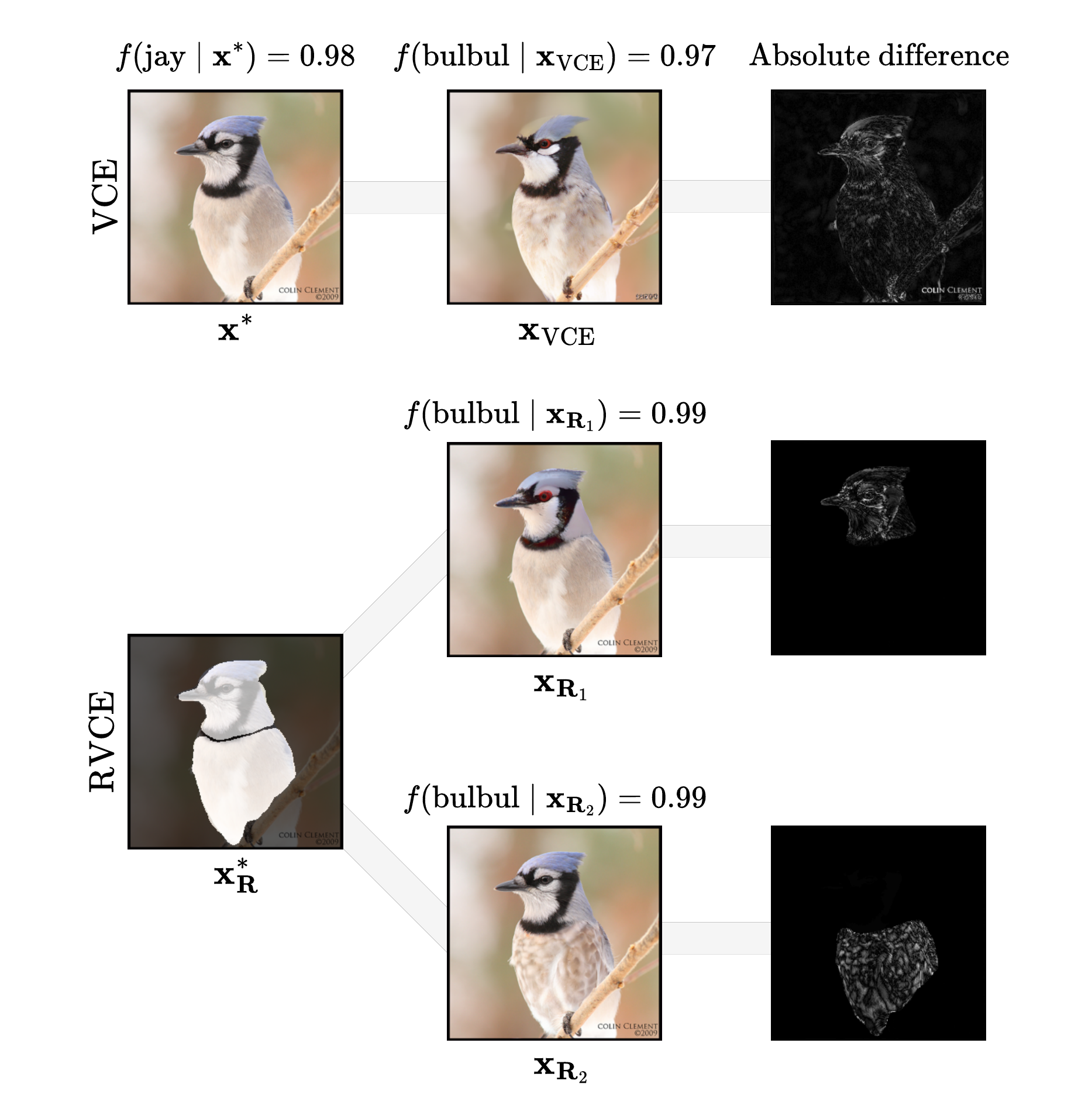}
    \caption{Previous methods create VCEs with unconstrained changes, making it virtually impossible to understand the decision-making process of a model. We propose \emph{region-constrained} VCEs, establishing a new paradigm for comprehensible and actionable explanatory process.}
    \label{fig:1_teaser}
    \vspace{-1em}
\end{wrapfigure}

Consider the image $\mathbf{x}^*$ in \cref{fig:1_teaser}, which the classifier $f$ correctly predicts to be a jay. In essence, VCEs focus on semantically editing $\mathbf{x}^*$ so that the prediction of $f$ changes to some target class -- bulbul in this case -- hence providing an answer to a specific \emph{what-if} question, through which the model's reasoning is explained. Consider now an example VCE for $\mathbf{x}^*$, denoted as $\mathbf{x}_{\text{VCE}}$, obtained with a recent state-of-the-art (SOTA) method. While $\mathbf{x}_{\text{VCE}}$ is successful at changing the prediction of $f$ and can be considered both realistic and semantically close to $\mathbf{x}^*$, answering \emph{why} $f$ now predicts it as a bulbul is close to impossible. 
The algorithm simultaneously modifies the bird's head and feathers, changes the texture of the branch and even modifies the copyright caption. The entanglement and dispersion of introduced changes hence leaves the question unanswered. We argue that to circumvent these fundamental difficulties, VCEs should be synthesized with a hard constraint on the \emph{region}, where the changes are allowed to appear, while leaving the rest of the image unchanged. For example, consider the image $\mathbf{x}_{\mathbf{R}}^*$ with regions of the bird's head ($\mathbf{R}_1$) and body ($\mathbf{R}_2$) overlayed. Constraining the VCEs to introduce changes \emph{only} to predetermined regions leads to two distinct explanations, $\mathbf{x}_{\mathbf{R}_1}$ and $\mathbf{x}_{\mathbf{R}_2}$, of why the decision changes to bulbul. By isolating the modified factors, the explanatory process greatly simplifies -- one can now state with certainty that $f$'s new prediction is based either on the modified feathers ($\mathbf{x}_{\mathbf{R}_2}$) or the changed characteristics of its head ($\mathbf{x}_{\mathbf{R}_1}$). Region-constrained VCEs (RVCEs) allow, therefore, to reason about the model's thought process in a \emph{causal} and principled manner, mitigating the potential \emph{confirmation bias} and clarifying the explanatory process.

By putting RVCEs in the spotlight, our work establishes new frontiers in the field of VCE generation. First, we define the objective of finding RVCEs as solving a conditional inpainting task. By building on top of the Image-to-Image Schrödinger Bridge (\isb{}, \cite{liu20232}) approach and adapting it to the classifier guidance scheme, we develop an efficient algorithm which synthesizes RVCEs with extreme realism, sparsity and closeness to the original image. Specifically, we set a new quantitative state-of-the-art (SOTA) on ImageNet \citep{deng2009imagenet} with up to 4 times better scores in FID and 3 times better sFID (realism), up to 2 times higher COUT (sparsity), and match or exceed S$^3$ (similarity) and Flip Rate (efficiency) achieved by previous methods. Through large-scale experiments, we demonstrate that, besides a fully automated way of synthesizing meaningful and highly interpretable RVCEs, our approach, \emph{Region-constrained Counterfactual Schrödinger Bridge} (\method{}), allows to infer causally about the model's change in prediction and enables the user to actively interact with the explanatory process by manually defining the region of interest. Moreover, our results highlight the importance of RVCEs in future research, indicating potential pitfalls of unconstrained methods that could lead to drawing misleading conclusions.

\section{Background \& Related Work}

In this section, we introduce the necessary background knowledge connected with score-based generative models (SGMs) and \isb{}, which forms the foundation of our method. We then present an overview of recent methods for VCE generation based on SGMs. For an extended literature review and detailed description of the theoretical basis, please refer to the \todo{Appendix}.

\textbf{SGM.} Following the work of \citet{songscore}, SGMs can be constructed through the framework of stochastic differential equations (SDEs), where samples from a complex distribution $p_0$ (\eg, natural images) are  mapped to a Gaussian distribution $p_1$, while the model is trained to reverse this mapping. Formally, converting data to noise is performed by following the \emph{forward} SDE (\cref{eq:main_forward_sde}), while denoising happens through the \emph{reverse} SDE (\cref{eq:main_reverse_sde}, \citet{anderson1982reverse}):
\begin{subequations}
\begin{align}
    \diff \mathbf{x}_t &= \mathbf{F}_t(\mathbf{\mathbf{x}_t}) \diff t + \sqrt{\beta_t} \diff \mathbf{w}, \label{eq:main_forward_sde} \\
    \diff \mathbf{x}_t &=(\mathbf{F}_t(\mathbf{\mathbf{x}_t}) - \beta_t \scoret )\diff t + \sqrt{\beta_t} \diff \mathbf{\Bar{w}},
    \label{eq:main_reverse_sde}
\end{align}
\end{subequations}
where $\mathbf{x}_t$ is the noisy version of a clean image $\mathbf{x} \in \mathbb{R}^n$ for some $n\in \mathbb{N}$ at timestep $t\in[0,1]$ , $\mathbf{w}$ and $\Bar{\mathbf{w}}$ denote the Wiener process and its reversed (in time) counterpart, $\mathbf{F}_t(\mathbf{x}_t): \mathbb{R}^n \rightarrow \mathbb{R}^n$ is the \emph{drift} coefficient, $\beta_t \in \mathbb{R}$ is the \emph{diffusion} coefficient and $\scoret$ is the \emph{score function}. An SGM $\mathbf{s}_{\boldsymbol{\theta}}$, where $\boldsymbol{\theta}$ denotes the model's parameters, is trained to approximate the score, \ie, $\mathbf{s}_{\boldsymbol{\theta}}(\mathbf{x}_t, t) \approx \scoret$. During sampling, denoising begins from pure noise $\mathbf{x}_1 \sim p_1$ and follows some discretized version of \cref{eq:main_reverse_sde} with the approximate score $\mathbf{s}_{\boldsymbol{\theta}}$. 

SGMs can also be adapted to \emph{conditional} generation, where $\mathbf{y}$ represents the conditioning variable. In this case, the score $\scoret$ is replaced by $\scoretwithcond$, which can be decomposed with Bayes' Theorem into $\scoretwithcond = \scoret + \scoretcond$. While $\scoret$ can be approximated with an already trained $\mathbf{s}_{\boldsymbol{\theta}}$, $\scoretcond$ must be modeled additionally. For $\mathbf{y}$ representing class labels, $p(\mathbf{y} \mid \mathbf{x}_t, t)$ can be approximated with an auxiliary time-dependent classifier $p_{\boldsymbol{\phi}}(\mathbf{y} \mid \mathbf{x}_t, t)$ trained on noisy images $\{ \mathbf{x}_t \}_{t\in [0,1]}$. Incorporating $p_{\boldsymbol{\phi}}$ into the sampling process is termed as \emph{classifier guidance} (CG), and can be strengthened (or weakened) with \emph{guidance scale} $s$ through $\scoret + s \cdot \scoretwithcond$. Therefore, class-conditional sampling in SGMs amounts to additionally maximizing the likelihood $p_{\boldsymbol{\phi}}(y \mid \mathbf{x}_t, t)$ of the classifier throughout the generative process to arrive at images from the data manifold, which resemble (according to $p_{\boldsymbol{\phi}}$) instances of a specific class. We emphasize this fact here for further reference.

\begin{wrapfigure}{r}{0.6\linewidth}
    \centering
    \includegraphics[width=1.0\linewidth]{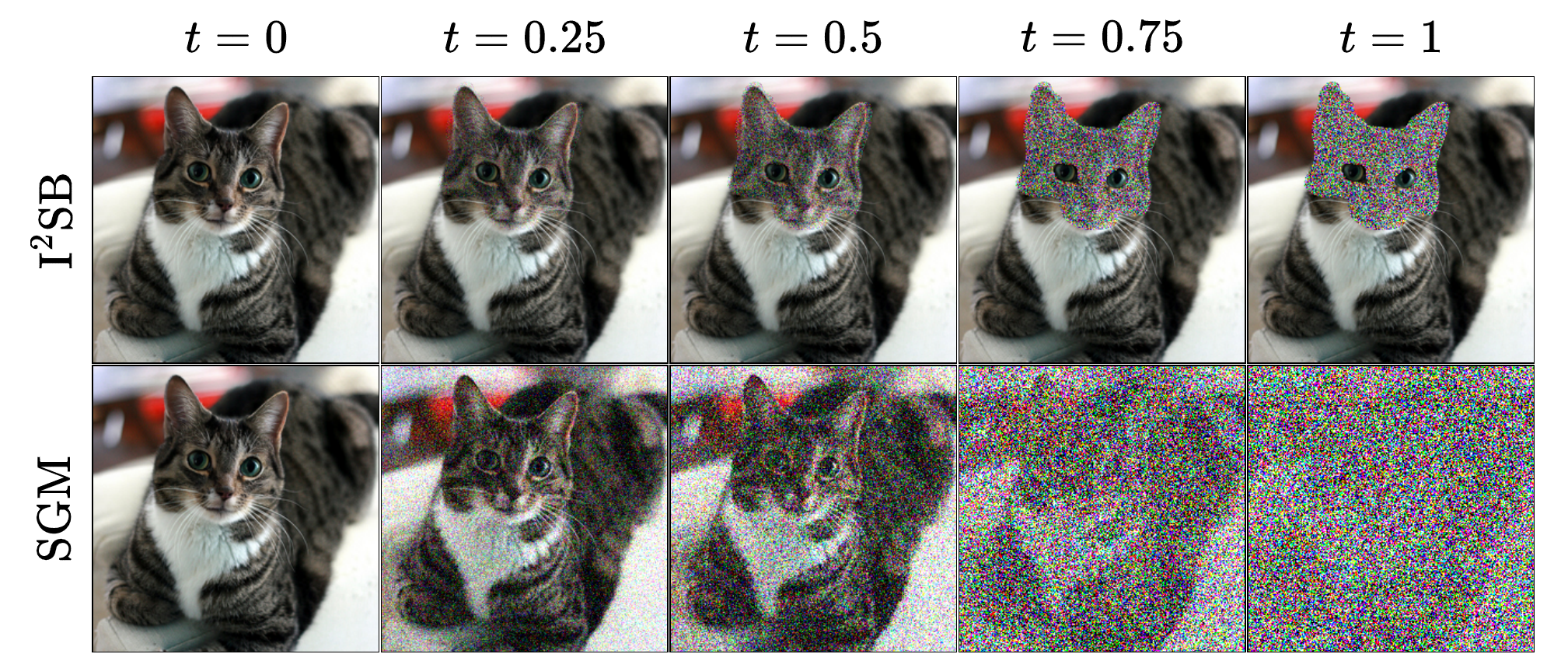}
    \vspace{-2em}
    \caption{Generative trajectories of \isb{} and SGM. Intermediate images of \isb{} are much closer to the data manifold.}
    \label{fig:3_trajs}
\end{wrapfigure}

\textbf{\isb{}.} The framework of \isb{} extends SGMs to $p_1$ representing an \emph{arbitrary} data distribution. For training, \isb{} requires paired data, \eg, in the form of clean and partially masked samples for inpainting, where it learns to infill the missing parts. While SGMs can also be adapted to solve inverse problems like inpainting, \isb{} maps these samples \emph{directly} (see \cref{fig:3_trajs} for a comparison of their generative trajectories). Therefore, \isb{} follows the same theoretical paradigm, where sampling is achieved by discretizing \cref{eq:main_reverse_sde} and using a score approximator $\mathbf{s}_{\boldsymbol{\psi}}$, but the generative process begins from a corrupted (\eg, masked) image instead of pure noise. Hence, \isb{} can also be adapted to conditional generation in the same manner as SGMs, especially for class-conditioning with an auxiliary classifier. Importantly, a special case of \isb{} follows an optimal transport ordinary differential equation (OT-ODE) when $\beta_t \rightarrow 0$, eliminating stochasticity beyond the initial sampling step (see \todo{Appendix}). We utilize the OT-ODE version of \isb{} in our implementation.

\textbf{SGM-based VCEs.} The initial approach of adapting SGMs to VCE generation, DiME \citep{jeanneret2022diffusion}, obtains the classifier's gradient by mapping the noised image to its clean version at each step through the reverse process. \citet{augustin2022diffusion} incorporate the gradient of a robust classifier and a cone projection scheme. \citet{jeanneret2023adversarial} decompose the VCE generation into pre-explanation construction and refinement using RePaint \citep{lugmayr2022repaint}. \citet{jeanneret2024text} utilize a foundation model, Stable Diffusion (SD, \citet{rombach2022high}), to generate VCEs in a black-box scenario. \citet{farid2023latent} and \citet{motzkus2024cola} utilize Latent Diffusion Models (LDMs), including SD, in a white-box context. \citet{weng2024fast} propose FastDiME to accelerate the generation process in a shortcut learning scenario. Also in black-box context, \citet{sobieski2024global} utilize a Diffusion Autoencoder \citep{preechakul2022diffusion} to find semantic latent directions that globally flip the classifier's decision. Finally, \citet{augustin2024digin} also make use of SD in various contexts, including classifier disagreement and neuron activation besides VCEs.

\section{Method}
In this section, we describe the details of our approach, beginning with the formulation of RVCEs as solutions to conditional inpainting task. Next, we motivate the use of \isb{} as an effective prior for synthesizing meaningful RVCEs and follow with a series of steps that better align the gradients of a standard classifier w.r.t corrupted images from its generative trajectory. We conclude with a description of the automated region extraction method, forming the basis of our algorithm.

\textbf{RVCEs through conditional inpainting.} We define the problem of finding RVCEs for the classifier $f$ from a given image $\mathbf{x}^*$, a region $\mathbf{R}$ and \emph{target class} label $y$, where $\argmax_{y'}f(y' \mid \mathbf{x}^*) \neq y$, as the task of sampling from 
\begin{equation}
    p(\mathbf{x} \mid \argmax_{y'}f(y' \mid \mathbf{x})=y, (\mathbf{1} - \mathbf{R}) \odot \mathbf{x} = (\mathbf{1} - \mathbf{R}) \odot \mathbf{x}^*), 
    \label{eq:rvces_distribution}
\end{equation}
where $\mathbf{R}$ is a binary mask with $1$ indicating the region. Intuitively, sampling from \cref{eq:rvces_distribution} means obtaining $\mathbf{x}$ with the complement of $\mathbf{R}$ unchanged and the content of $\mathbf{R}$ modified in a way that changes the decision of $f$ to $y$, \ie, performing inpainting with additional condition coming from the classifier $f$.

\begin{figure}
    \centering
    \includegraphics[width=1.0\linewidth]{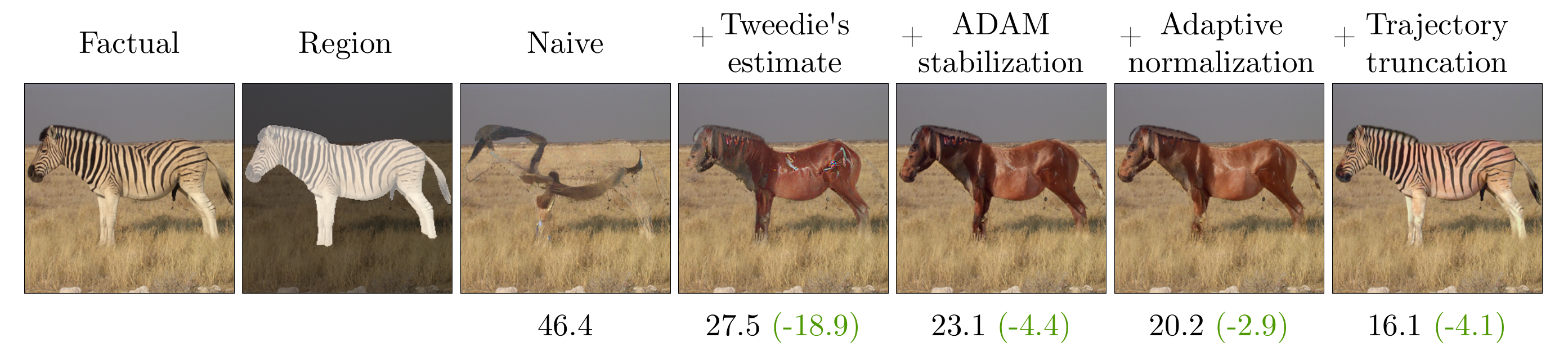}
    \vspace{-2em}
    \caption{Series of proposed improvements to better align the gradient's of the classifier of interest with the generative trajectory. Changes to the \emph{factual} image are constrained to the indicated \emph{region}. Subsequent images illustrate the influence of each new adaptation. Numbers below images correspond to FID ($\downarrow$) values obtained in a larger-scale experiment (for details, see \todo{Appendix}).}
    \label{fig:3_inc_imp}
    \vspace{-1em}
\end{figure}

\textbf{Synthesizing meaningful RVCEs.} Looking at \cref{eq:rvces_distribution}, one quickly realizes that obtaining semantically meaningful RVCEs requires maximizing the likelihood $f(y \mid \mathbf{x})$ of the classifier while inpainting $\mathbf{R}$ with content that keeps $\mathbf{x}$ in the data manifold. These conditions greatly resemble the CG scheme in the context of \isb{}, since the score estimate $\boldsymbol{s}_{\boldsymbol{\psi}}$ serves as an effective prior for generating in-manifold infills, while the likelihood $p_{\boldsymbol{\phi}}(y \mid \mathbf{x})$ of an auxiliary classifier is maximized to ensure that $p_{\boldsymbol{\phi}}$ predicts them as instances of $y$. Moreover, \isb{} maps masked images \emph{directly} to clean samples, leaving the content outside $\mathbf{R}$ unchanged in the final image.

The above arguments suggest that inserting $f$ in place of $p_{\boldsymbol{\phi}}$ should function as an effective mechanism for sampling meaningful RVCEs. However, a fundamental drawback of this \emph{naive} approach is that, throughout the generative process, $f$'s gradients originate from evaluating it on images with highly noised infills inside $\mathbf{R}$ (see \cref{fig:3_trajs}). Such corrupted images are far from what $f$ observed during training, hence leading to a \emph{misalignment} of its gradients with the correct trajectory and generation of out-of-manifold samples. Similar issue has been identified by previously mentioned SGM-based methods for VCEs, which can be generally unified as attempts to replace the auxiliary classifier $p_{\boldsymbol{\phi}}$ with $f$ in the CG scheme in SGMs and \emph{correct} $f$'s gradients. Following \cref{fig:3_trajs}, one should expect the misalignment in these methods to be of great extent, as the generative trajectory consists of highly noised images, leaving no meaningful content for $f$ to provide accurate gradients. There, as shown in \cref{fig:3_trajs}, \isb{} provides a crucial advantage, which stems from its generative trajectory being \emph{much closer} to the data manifold. Moreover, by using \isb{}, $f$ is able to effectively utilize the readily available context outside $\mathbf{R}$.
Hence, in the following, we focus on reducing the misalignment problem caused by the noised content inside $\mathbf{R}$, in the end arriving at a highly effective algorithm for meaningful RVCEs.

\textbf{Aligning the gradients.} We propose to adapt the gradients of $f$ to properly align with the generative trajectory of \isb{} through a series of incremental steps. To provide the intuition standing behind the introduction of each consecutive improvement, \cref{fig:3_inc_imp} provides an example RVCE task, where the factual image depicts a zebra correctly predicted by the model (ResNet50 \citep{he2016deep}), and the goal is to change the decision to `sorrel'. We set the region constraint to include the entire animal to make the task challenging enough and verify the improvements quantitatively through a large-scale experiment with around $2000$ images. For each step, we compute FID between the RVCEs and original images to assess their realism. For details on the experimental setup, \todo{see Appendix.}

\textbf{Naive.} We first verify that naively plugging $f$ in place of $p_{\boldsymbol{\phi}}$ does not provide meaningful results. Indeed, as shown in \cref{fig:3_inc_imp}, the method struggles to include the information from $f$. The unrealistic infill also suggests that the classifier's signal negatively influences the score from \isb{}.

\textbf{Tweedie's formula.} To begin with closing the gap between the data manifold and the generative trajectory, we refer to a classic result of Tweedie's formula \citep{robbins1992empirical, chung2022improving}, which states that a \emph{denoised estimate} of the final image at step $t$ can be achieved by computing the posterior expectation
\begin{equation}
    \hat{\mathbf{x}}_0(\mathbf{x}_t) := \mathbb{E}[\mathbf{x}_0 \mid \mathbf{x}_t] = \mathbf{x}_t + \sigma^2_t \scoret,
    \label{eqn:tweedie}
\end{equation}
where $\sigma^2_t = \int_0^t \beta_\tau \diff \tau$. For visual differences between $\mathbf{x}_t$ and $\hat{\mathbf{x}}_0(\mathbf{x}_t)$, see \todo{Appendix}. Crucially, one has access to approximate $\hat{\mathbf{x}}_0(\mathbf{x}_t)$ at every step $t$ by utilizing \isb{} as the approximate score. Replacing $\nabla_{\mathbf{x}_t}\log{f(y \mid \mathbf{x}_t)}$ with $\nabla_{\mathbf{x}_t}\log{f(y \mid \hat{\mathbf{x}}_0 (\mathbf{x}_t))}$ brings the inputs of $f$ much closer to what it expects, improving the conditional inpainting process as indicated by \cref{fig:3_inc_imp}, which now shows a structure resembling a sorrel and a much smaller FID.

\textbf{ADAM stabilization.} Despite utilizing the Tweedie's estimate, we observed the norms of $f$'s gradients to have a very noisy tendency throughout the generation process, pointing out a possible cause for visible artifacts and the missing parts of the animal. Hence, we propose to smooth out the gradients by applying the ADAM update rule at each step \citep{vaeth2024gradcheck,kingmaadam}, to which we simply refer as ADAM stabilization. \Cref{fig:3_inc_imp} indicates that this modification allows for filling in the missing parts of the sorrel and further lowering FID.

\textbf{Adaptive normalization.} Incorporating ADAM stabilization required greatly lowering the guidance scale to values on the order of $1\mathrm{e}{-2}$, as using standard $s=1$ led to extreme artifacts. This phenomenon suggested that the step size could also be adjusted throughout the generation process. While we initially experimented with various types of schedulers (see \todo{Appendix}), using \emph{adaptive normalization} has empirically proven to be the most effective approach. Specifically, at the beginning of the conditional inpainting process, we register the norm of the first encountered gradient of the log-likelihood of $f$. We then use it as a normalizing constant for each subsequent gradient, meaning that the generation begins with gradient of unit norm. This simple modification not only further lowered FID, but also reduced the final visible artifacts and improved color balance (\cref{fig:3_inc_imp}).

\textbf{Trajectory truncation.} Up until this point, we relied solely on the ability of \isb{} and the classifier's signal to correctly infill the missing regions with semantically meaningful content, with no knowledge of the structure of the missing objects. Since a possible infill of the region is always available from the original image, one can begin the inpainting process from some \emph{intermediate} step instead of the final one. This intervention allows for mixing the available information with the one coming from the classifier, and gives direct control over the preservation of the original content. As our approach does not bias the conditional score with signal from any additional losses (like Learned Perceptual Image Patch Similarity (LPIPS \citet{zhang2018unreasonable}) or $l_2$ in other works), we can fully rely on the \emph{conceptual} compression of \isb{}, similarly to SGMs \citep{ho2020denoising}, which decomposes the generation process into initial phases responsible for the overall structure of objects and later ones responsible for small details. \Cref{fig:3_inc_imp} showcases the effect of using this \emph{trajectory truncation} ($\tau$) at the $0.4$ level, meaning that the infilling process starts from $t=\tau \cdot T$, where $T$ denotes the final timestep. 
\begin{wrapfigure}{r}{0.5\linewidth}
    \vspace{-1em}
    \includegraphics[width=\linewidth]{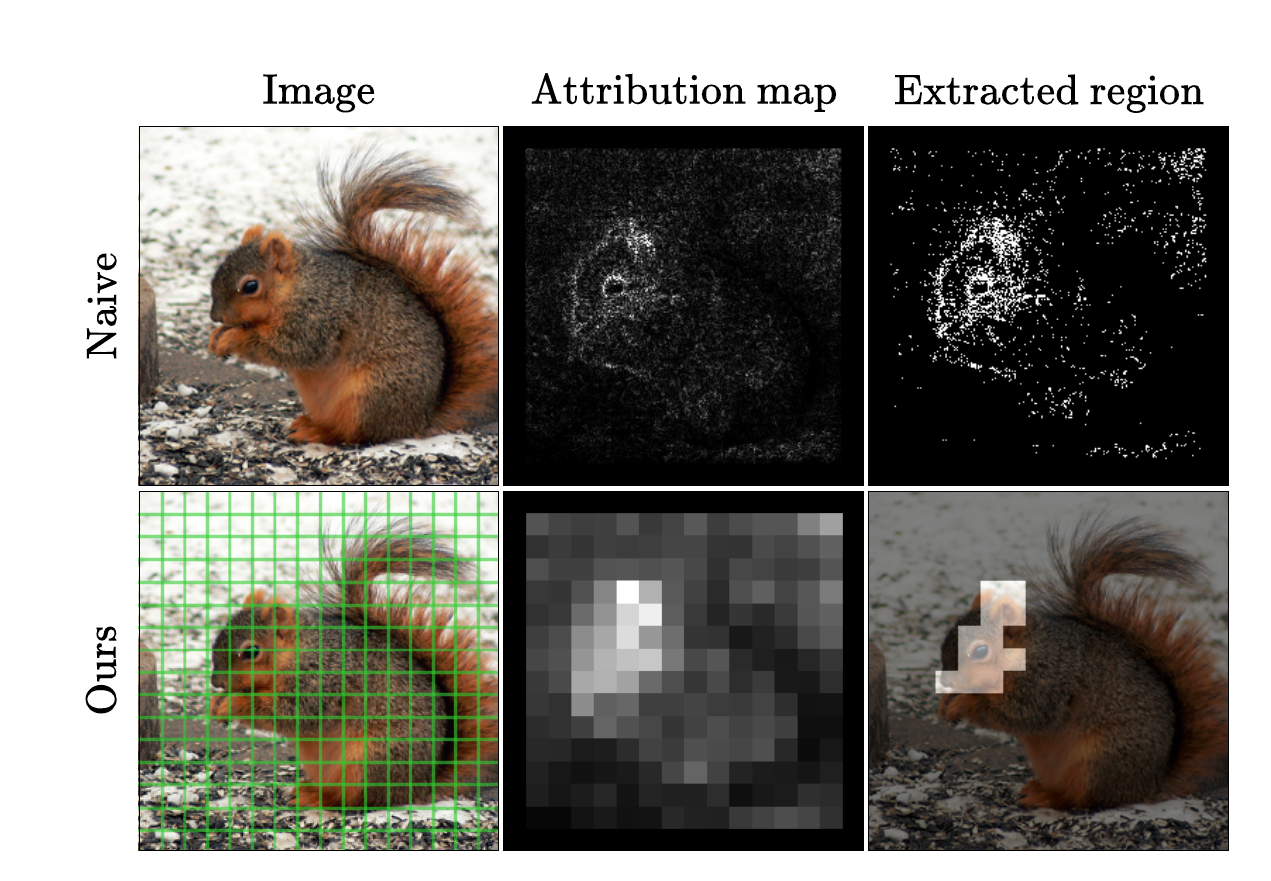}
    \caption{Example region obtained with our automated region extraction. Instead of directly binarizing an attribution map (upper row), we amplify the focus on semantic concepts (bottom row) with a simple approach based on grid cells.}
    \label{fig:3_reg_extr}
    \vspace{-1em}
\end{wrapfigure}
Understandably, trajectory truncation greatly lowers the FID score, as much more information is available from the very beginning of the process, and introduces much more subtle changes to the image. We explore the effect of manipulating $\tau$ further \todo{in the Appendix}, showing that it functions as a very interpretable mechanism for controlling the content preservation.

\textbf{Automated region extraction.} While the introduced algorithmical improvements effectively incorporate the classifier's signal into the inpainting process, they do not address the issue of predetermining the region for the resulting explanation. To this end, the optimal strategy would be fully automated and focus on regions that are both important to the classifier's prediction and point to semantically meaningful concepts. This description closely resembles the role of visual attribution methods, which assign importance values to pixels based on their relevance to the model's output \citep{holzinger2022explainable}. \Cref{fig:3_reg_extr} shows an example attribution map obtained with Integrated Gradients (IG, \citet{sundararajan2017axiomatic}) method for the squirrel prediction of a ResNet50 model. Perceptually, highest attributions are focused around the squirrel's head. To extract a region from such attributions, one can threshold them to cover a specific fraction $a$ of the total image area. However, after binarizing the attributions with $a=0.05$, we observe that the resulting region is highly scattered, losing focus from semantic concepts. To address this issue, we divide the image into a grid of square cells of size $c \times c$, where each cell receives the value equal to the sum of the absolute pixel attributions inside it. \Cref{fig:3_reg_extr} shows that this postprocessing mechanism (here with $c=16$) greatly amplifies the focus of the resulting map. By thresholding it with $a=0.05$, we observe the extracted region to focus solely on the squirrel's head. This leads to a fully automated strategy for obtaining regions that are both aligned with semantically meaningful concepts and based on pixels that are important for the classifier.

We term the final version of the algorithm which combines all of the aforementioned improvements with the automated region extraction as \method{}. For the pseudocode of the entire procedure, see Appendix. We include our implementation at \href{https://github.com/sobieskibj/rcsb}{https://github.com/sobieskibj/rcsb}.

\section{Experiments}

\begin{wraptable}{r}{0.5\linewidth}
\vspace{-1em}
\centering
    \scriptsize
    \begin{tabular}{c|*{5}{c}}
    \toprule
    Method & FID & sFID & S$^3$ & COUT & FR \\
    \midrule
    \multicolumn{6}{c}{\textbf{Zebra -- Sorrel}} \\
    \midrule
    ACE $l_1$ & $84.5$ & $122.7$ & $\mathbf{0.92}$ & $-0.45$ & $47.0$ \\
    ACE $l_2$ & $67.7$ & $98.4$ & $\underline{0.90}$ & $-0.25$ & $81.0$ \\
    LDCE-cls & $84.2$ & $107.2$ & $0.78$ & $-0.06$ & $88.0$\\
    LDCE-txt & $82.4$ & $107.2$ & $0.71$ & $-0.21$ & $81.0$ \\
    DVCE & $33.1$ & $43.9$ & $0.62$ & $-0.21$ & $57.8$ \\
    \method{}$^C$& $13.0$ & $20.4$ & $0.82$ & $0.70$ & $\mathbf{99.7}$\\
    \method{}$^B$ & $\underline{9.51}$ & $\underline{17.4}$ & $0.86$ & $\underline{0.72}$ & $97.4$\\
    \method{}$^A$ & $\mathbf{8.0}$ & $\mathbf{16.2}$ & $0.88$ & $\mathbf{0.74}$ & $\underline{94.7}$\\
    \midrule
    \multicolumn{6}{c}{\textbf{Cheetah -- Cougar}} \\
    \midrule
    ACE $l_1$ & $70.2$ & $100.5$ & $\underline{0.91}$ & $0.02$ & $77.0$\\
    ACE $l_2$ & $74.1$ & $102.5$ & $0.88$ & $0.12$ & $95.0$\\
    LDCE-cls & $71.0$ & $91.8$ & $0.62$ & $0.51$ & $\underline{100.0}$\\
    LDCE-txt & $91.2$ & $117.0$ & $0.59$ & $0.34$ & $98.0$\\
    DVCE & $46.9$ & $54.1$ & $0.70$ & $0.49$ & $99.0$ \\
    \method{}$^C$ & $30.2$ & $39.2$ & $0.87$ & $0.79$ & $\underline{100.0}$\\
    \method{}$^B$ & $\underline{23.4}$ & $\underline{32.4}$ & $0.90$ & $\underline{0.85}$ & $99.9$\\
    \method{}$^A$ & $\mathbf{17.2}$ & $\mathbf{26.6}$ & $\mathbf{0.92}$ & $\mathbf{0.92}$ & $\mathbf{100.0}$\\
    \midrule
    \multicolumn{6}{c}{\textbf{Egyptian Cat – Persian Cat}} \\
    \midrule
    ACE $l_1$ & $93.6$ & $156.7$ & $\underline{0.85}$ & $0.25$ & $85.0$\\
    ACE $l_2$ & $107.3$ & $160.4$ & $0.78$ & $0.34$ & $97.0$ \\
    LDCE-cls & $102.7$ & $140.7$ & $0.63$ & $0.52$ & $99.0$ \\
    LDCE-txt & $121.7$ & $162.4$ & $0.61$ & $0.56$ & $99.0$ \\
    DVCE & $46.6$ & $59.2$ & $0.59$ & $0.60$ & $98.5$ \\
    \method{}$^C$ & $41.1$ & $56.3$ & $0.79$ & $0.82$ & $\underline{100.0}$\\
    \method{}$^B$ & $\underline{31.3}$ & $\underline{48.1}$ & $0.84$ & $\underline{0.87}$ & $\underline{100.0}$\\
    \method{}$^A$ & $\mathbf{23.0}$ & $\mathbf{40.0}$ & $\mathbf{0.87}$ & $\mathbf{0.92}$ & $\mathbf{100.0}$\\
    \bottomrule
    \end{tabular}
    \caption{Quantitative comparison with SOTA. \method{} outperforms previous methods by a large margin across all metrics. The best results are obtained with $A(a=0.1,c=4,s=3,\tau=0.6)$, but the superiority is clear for various configurations, including $B(a=0.2,c=4,s=1.5,\tau = 0.6)$, $C(a=0.3,c=4,s=1.5,\tau=0.6).$}
    \label{tab:main_results}
    \vspace{-1em}
\end{wraptable}
Following previous works for VCEs on ImageNet, we base the quantitative evaluation on 3 challenging \textbf{main} VCE generation tasks: \textbf{Zebra -- Sorrel}, \textbf{Cheetah -- Cougar}, \textbf{Egyptian Cat – Persian Cat}, where each task requires creating VCEs for images from both classes and flipping the decision to their counterparts. We treat it as a general benchmark for evaluating the effectiveness of \method{} in various scenarios. We use FID ($\downarrow$) and sFID ($\downarrow$) to assess realism \citep{heusel2017gans}, S$^3$ ($\uparrow$) for representation similarity \citep{chen2021exploring}, COUT $\in [-1, 1]\ (\uparrow)$ \citep{khorram2022cycle} for sparsity and Flip Rate (FR) ($\uparrow$) for efficiency. 
For qualitative examples, we extend the main tasks with a large array of \textbf{other} tasks, which we show throughout the paper and the Appendix, where more details regarding the experimental setup and the metrics description can be found.

\textbf{RCSB sets new SOTA for VCEs.} We first verify that synthesizing RVCEs with \method{} leads to new SOTA in VCE generation. \cref{tab:main_results} quantitatively compares \method{} with recent SOTA approaches to VCEs on ImageNet. Our RVCEs are much more realistic (at least $2-4\times$ decrease in FID and sFID), stay close to original images (match or exceed best values of S$^3$) and almost always flip the model's decision ($\text{FR}\approx1.0$). \method{} also solves a long-standing challenge of achieving extremely \emph{sparse} explanations on ImageNet, especially on \textbf{Zebra -- Sorrel} task. While all other methods fail to achieve nonnegative values, \method{} approaches the upper bound of COUT. Our method is clearly the most balanced, as it does not struggle on any specific metric like, \eg, DVCE on S$^3$. In the Appendix, we show that it is also the most computationally efficient.

\begin{figure}
    \centering
    \includegraphics[width=1.0\linewidth]{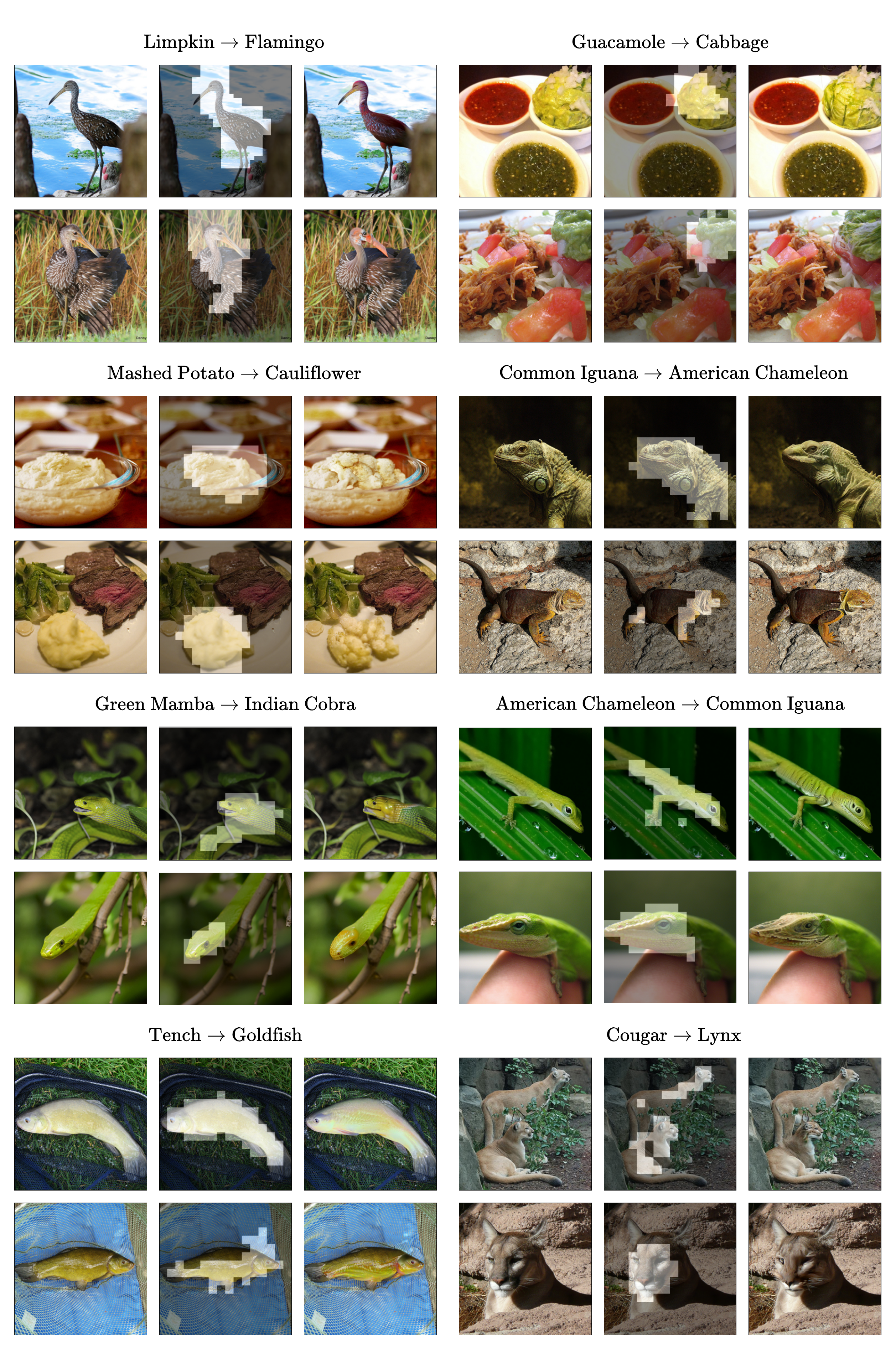}
    \vspace{-2em}
    \caption{Qualitative examples obtained with \method{} using automated region extraction. Each task of the form \emph{predicted class} $\rightarrow$ \emph{target class} shows the factual image, the extracted region and the RVCE obtained with \method{}.}
    \label{fig:4_main_automated}
\end{figure}

\Cref{fig:4_main_automated} shows example explanations obtained with \method{}, greatly highlighting the importance of synthesizing RVCEs instead of standard VCEs. Our region extraction approach is able to precisely localize semantic concepts responsible for the model's decision. For example, in the Guacamole $\rightarrow$ Cabbage task, \method{} detects the guacamole bowl in the background and, guided by the classifier, infills it with cabbage while leaving the rest of the image unchanged. \method{} is capable of performing a wide range of editing tasks with various levels of difficulty, beginning with textural and color-based edits (\eg, Tench $\rightarrow$ Goldfish, Mashed Potato $\rightarrow$ Cauliflower) to partially changing the object's structure (\eg, Limpkin $\rightarrow$ Flamingo) to infilling the region with new, realistically looking concepts (\eg, Cougar $\rightarrow$ Lynx, Green Mamba $\rightarrow$ Indian Cobra, Cougar $\rightarrow$ Lynx). Most importantly, thanks to the region constraint, our RVCEs allow for greatly limiting the potential factors that influenced the model's decision, making the explanations much more interpretable.

\textbf{RCSB allows for causal inference about the model's reasoning.} Drawing definite conclusions about the model's reasoning from an unconstrained VCE is not possible, as one cannot be certain that modifying potentially irrelevant factors did not in fact influence the prediction. RVCEs overcome this limitation when constrained on the region connected with the sole factor of interest, \eg, the body of an animal in a species prediction task. To adapt \method{} to such scenario, we replace the automated region extraction method with a foundation text-to-object-segmentation model \footnote{Language Segment Anything (\href{https://github.com/paulguerrero/lang-sam}{LangSAM}) combines Segment Anything Model \citep{kirillov2023segment} with GroundingDINO \citep{liu2024grounding} to allow object segmentation from text prompts.}. Using the class name from a given task as the text prompt allows us to obtain highly precise segmentation masks of the relevant objects, enabling the identification of the cause behind the model's prediction change based solely on factors related to the object of interest.

We first quantitatively assess that \method{} is capable of utilizing regions provided by a generic object detector at scale. \Cref{tab:additional_results}{}(A) shows the results of this evaluation together with the used text prompt. Here, the metrics are computed by first discarding images with a mask that covers area larger than $40\%$. Despite \isb{} being trained on masks covering at most $30\%$ of the image area, we observed that it generalizes well beyond this threshold with $40\%$ starting to pose a challenge. Crucially, despite the regions being classifier-agnostic and hence not necessarily focused on the most influential pixels, \cref{tab:additional_results}(A) indicates that \method{} is versatile enough to maintain most of the performance from the automated approach. The efficiency, sparsity and representation similarity of the obtained RVCEs remain very close to the values achieved by the closest configuration (in terms of hyperparameters) from \cref{tab:main_results}, as the region area is often close to or exceeds $30\%$. The slight increase in FID and sFID stems mainly from the regions covering complex objects, whose modification may naturally move RVCEs further from original data at a distribution level, and a lower number of images used for these metrics' computation (as both are sensitive to sample size) due to the rejection of samples from the area constraint.

\begin{table}
\addtolength{\tabcolsep}{-0.2em}
\centering
    \scriptsize
    \begin{tabular}{l|*{5}{c}|*{5}{c}|*{5}{c}}
    \toprule
   Metric & FID & sFID & S$^3$ & COUT & FR & FID & sFID & S$^3$ & COUT & FR & FID & sFID & S$^3$ & COUT & FR \\
    \midrule
    Task & \multicolumn{5}{c}{\textbf{Zebra -- Sorrel}} & \multicolumn{5}{|c|}{\textbf{Cheetah -- Cougar}} & \multicolumn{5}{c}{\textbf{Egyptian Cat – Persian Cat}} \\
    \midrule
    A & \multicolumn{15}{c}{Exact regions obtained with LangSAM and prompts: zebra / horse, cheetah / cougar, cat respectively} \\
    \midrule
    Values & 32.8 & 41.5  & 0.87 & 0.74 & 98.9 & 37.2 & 50.6 & 0.91 & 0.84 & 99.4 & 52.0 & 82.8 & 0.81 & 0.84 & 99.2\\
    \midrule
    B & \multicolumn{15}{c}{Regions based on freeform masks with the area in the indicated range}  \\
    \midrule
    $10-20\%$  & 6.7 & 15.0 & 0.85 & 0.85 & 87.6 & 9.0 & 19.1 & 0.89 & 0.72 & 96.6 & 12.4 & 29.6 & 0.80 & 0.73 & 96.9 \\
    $20-30\%$  &  7.8 & 15.8 & 0.84 & 0.53 & 92.2 & 11.6 & 21.3 & 0.88 & 0.71 & 99.6 & 17.7 & 34.0 & 0.78 & 0.74 & 99.3 \\
    \midrule
    C & \multicolumn{15}{c}{Ablation study with adaptations of other inpainting algorithms}  \\
     \midrule
    RePaint & 63.8 & 76.0 & 0.55 & 0.77 & 99.3 & 129.3 & 144.2 & 0.50 & 0.77 & 99.0  & 148.7 & 175.2 & 0.38 & 0.76 & 99.5 \\
    MCG & 43.2 & 55.6 & 0.73 & 0.45 & 96.0 & 76.6 & 91.4 & 0.74 & 0.64 & 100.0 & 93.7 & 117.5 & 0.62 & 0.65 & 99.9 \\
    DDRM & 42.5 & 49.4 & 0.69 & 0.72 & 99.6 & 60.5 & 68.4 & 0.72 & 0.76 & 100.0 & 59.2 & 73.0 & 0.63 & 0.76 & 100.0 \\
    \bottomrule
    \end{tabular}
    \caption{Quantitative results from various experiments. \textbf{A}:~regions extracted from LangSAM with text prompt connected to the initial class name. \textbf{B}:~regions based on freeform masks that cover the fraction of the total area from the indicated range. \textbf{C}:~automatically extracted regions used with adaptations of other inpainting algorithms.}
    \label{tab:additional_results}
    \vspace{-2em}
\end{table}

\begin{figure}
    \centering
    \includegraphics[width=1.0\linewidth]{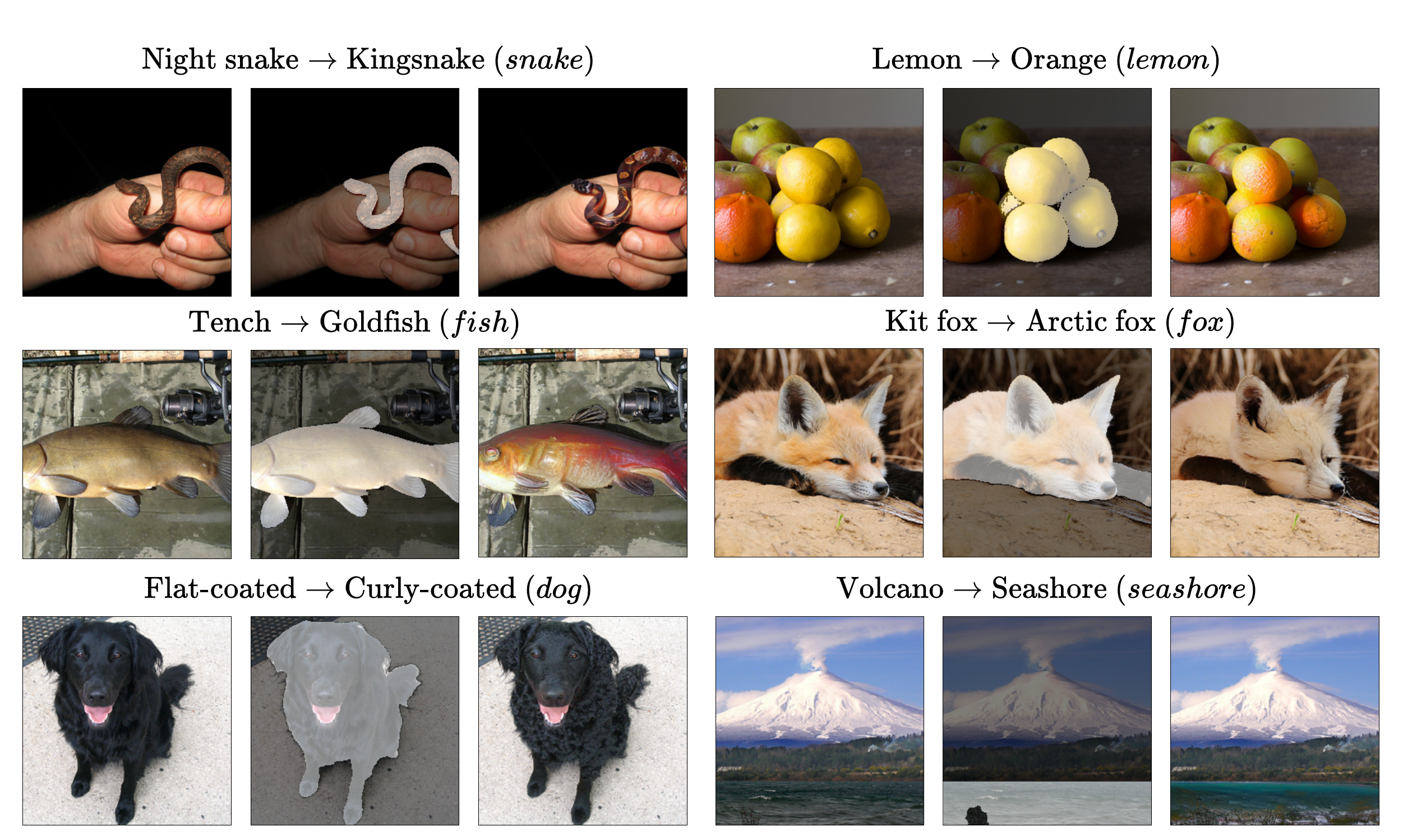}
    \vspace{-2em}
    \caption{Qualitative examples obtained with \method{} using \emph{exact} regions extracted from LangSAM using text prompt of the predicted class. For each task of the form \emph{predicted class} $\rightarrow$ \emph{target class}, a factual image together with the used region and the resulting RVCE are shown. The used text prompts are \emph{emphasized}.}
    \label{fig:4_main_lang_sam}
\end{figure}

Regions that contain \emph{exactly} the objects of interest provide novel insights about the model's reasoning. For example, consider the Lemon $\rightarrow$ Orange task from \cref{fig:4_main_lang_sam},  where the lemons were correctly identified by the ResNet50 model. One would require the VCE for this task to indicate the sole determining factor of 'why lemons and not oranges'. However, with unconstrained VCEs, this identification process quickly becomes incomprehensible due to small changes added to each object in the image, such as other fruits. By constraining VCEs to the region occupied by the lemons, the reasoning process can be disentangled and simplified, as one can now look for this factor in the modifications of the lemons only. In this case, \method{} allows for increasing trust in the model, as making the lemons more orange correctly modifies its decision. 

RVCEs also allow for clarifying the model's decision-making when its reasoning is not initially understandable. In the Volcano $\rightarrow$ Seashore task, the image shows both objects, while the model predicts it as the former. Applying \method{} to the exact region of the seashore results in a RVCE that changes the model's decision when the water's color becomes more light blue and structures like stones start to appear. Hence, one is able to better understand what the model \emph{actually} identifies as a seashore. In other examples, the method introduces class-specific characteristics when the changes are constrained \emph{precisely} and \emph{exclusively} to the object of interest, ensuring the receiver about the general cause of the model's decision change. Such cases are also especially relevant when the generative model used to synthesize explanations is prone to systematic errors like, \eg, SGMs struggling with correctly generating hands. In the Night Snake $\rightarrow$ Kingsnake task, this error can be bypassed with the region constraint by not allowing the generative model to affect anything other than the animal, hence alleviating the evaluation of the classifier on out-of-manifold samples.

\textbf{Discovering complex patterns with interactive RVCEs.} Despite the impressive capabilities of deep models in object localization, the receiver of the explanation may be interested in testing the model for highly abstract and complex concepts that cannot be localized automatically and must be provided manually by the user. We begin with verifying the capability of \method{} in generating RVCEs based on user-defined regions by simulating such scenario at scale. Specifically, we randomly match images from the main tasks with regions given by the $10\%-20\%$ and $20\%-30\%$ freeform masks from the \isb{} training data \citep{saharia2022palette}. We argue that this serves as a very challenging benchmark, since the algorithm's access to the most influential pixels (for the classifier) might often be very restricted.

Despite the task's difficulty, quantitative results from \cref{tab:additional_results}(B) highlight the versatility of \method{}, which is able to effectively utilize the restricted resources to influence the classifier's prediction. While S$^3$, COUT and FR are not significantly different from previous results, we observe a decrease in FID and sFID, indicating higher realism and closeness to the data distribution. This is largely due to the fact that freeform masks are often not connected to entire complex objects and do not contain the pixels most important to the classifier. Hence, \method{} may often leave large portions of the regions unchanged, which boosts the realism evaluation.

\begin{figure}[t]
    \centering
    \includegraphics[width=1.0\linewidth]{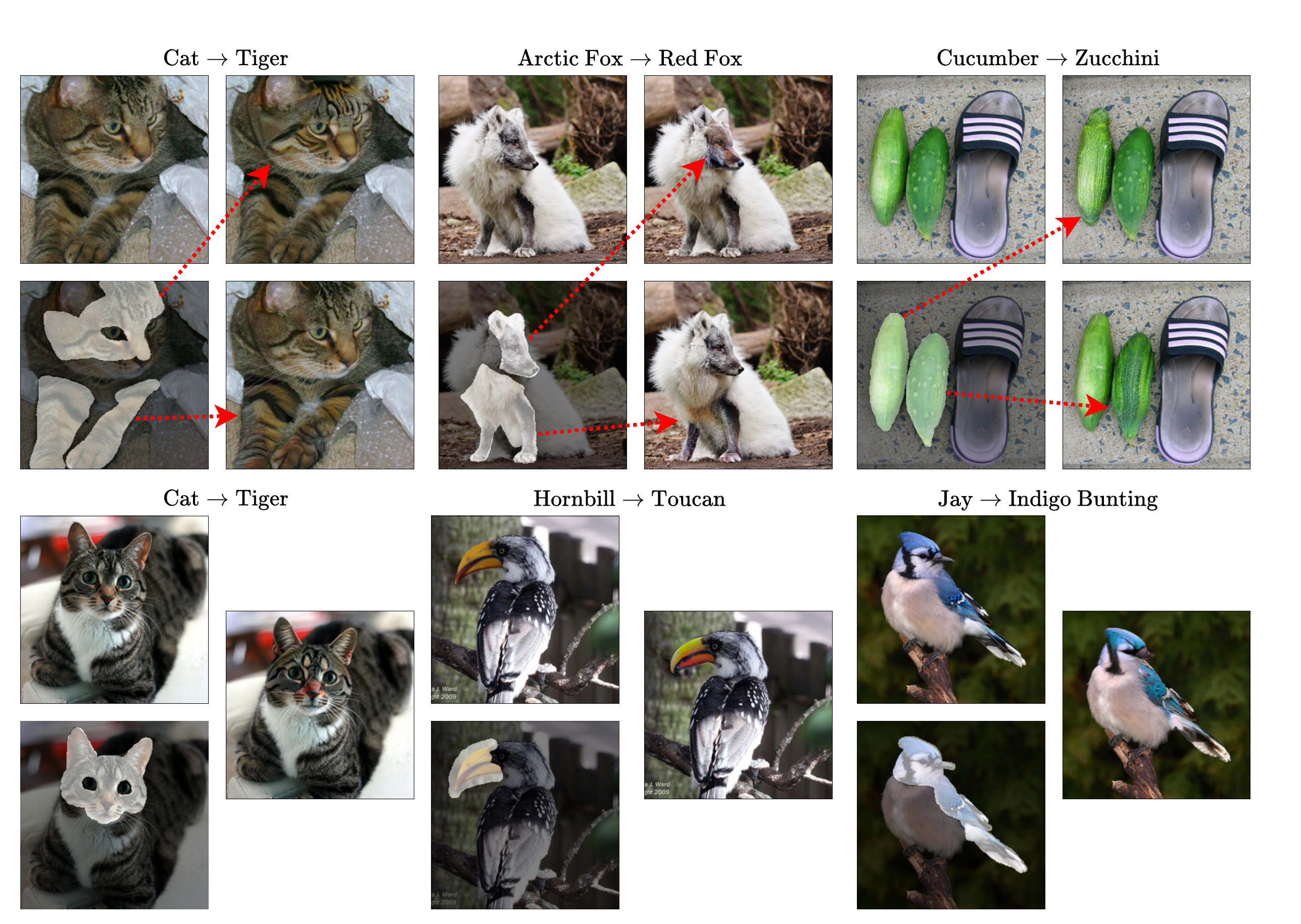}
    \caption{Qualitative examples obtained with \method{} from \emph{user-defined} regions. For each task of the form \emph{predicted class} $\rightarrow$ \emph{target class}, a factual image together with the provided regions are shown. Arrows point to RVCEs obtained by modifying only the indicated region.}
    \label{fig:4_main_user_defined}
    \vspace{-2em}
\end{figure}

To allow for true interaction of the user with the explanatory process, we implement a simple interface that allows for manual image segmentation using a brush-like cursor. \Cref{fig:4_main_user_defined} shows example results, where we manually predefine the regions on different images. This exploration gives important insights about the added value provided by RVCEs. In the Cat $\rightarrow$ Tiger task, we discover that the classifier's decision can be flipped by independently modifying either the cat's paws or snout, in both cases introducing a tiger's coloration. Similarly in the Arctic Fox $\rightarrow$ Red Fox task, choosing either the ears and muzzle or paws and stomach area allows for changing the model's decision with the features of a red fox. User-defined regions also allow to discover unusual reasoning patterns of the model. In the Cucumber $\rightarrow$ Zucchini task, the model's decision can be influenced by modifying only one of the cucumbers to zucchini, leaving the other unchanged. This observation connects with recent positions on the topic of contextual and spatial understanding of predictive models \citep{tomaszewska2024position}, providing new rationale in further exploring how image classifiers \emph{actually} reason.

\textbf{Ablating \method{}'s components.} We empirically verified that combining our novel guidance mechanism with the \isb{} prior leads to highly effective RVCEs. To better understand the benefits provided by each component of our framework, we perform an ablation study, where we adapt the proposed improvements to SGM-based inpainters, aiming to assess the influence of the guidance scheme and \isb{} in isolation. Specifically, we pick RePaint \citep{lugmayr2022repaint}, one of the first adaptations of SGMs to inpainting, MCG \citep{chung2022improving} and DDRM \citep{kawar2022denoising}, two different adaptations of SGMs to linear inverse problems, which also include inpainting. We manually tune our guidance scheme to each method on a small subset of images and repeat the same evaluation protocol with the automated region extraction method (see Appendix for details of each adaptation). As these methods are much less compute-efficient, we cap their computational budget on each task to 24 A100 GPU hours.

\Cref{tab:additional_results}(C) shows the results of the ablation study. Despite the fact that the used methods were never explicitly trained for inpainting, combining them with our guidance mechanism and region extraction allows for matching or even exceeding previous SOTA. For example, all adaptations achieve very high sparsity, almost always flip the classifier's decision and keep the explanation close to the original. This indicates the benefits of utilizing only the pixels from the extracted region and a proper utilization of the classifier's gradients without biasing them with additional components like LPIPS or $l_2$ loss. \method{} differentiates itself from the adaptations with a much higher  realism of the obtained RVCEs (significantly lower FID and sFID), more balanced results and much smaller computational burden, \eg $24\times$ less NFEs than RePaint. These benefits stem from the \isb{} prior, which is trained to map corrupted images directly to clean samples and the resulting trajectory being much closer to the data manifold, allowing the classifier to more effectively influence the inpainting process.

For extended quantitative and qualitative results, including RVCEs obtained for another 5 non-robust, 2 robust and a zero-shot CLIP classifier \citep{radford2021learning}, and evaluation of 10 other attribution methods, we refer to the Appendix.

\section{Conclusions}

Our work advances the SOTA in VCE generation by constraining the explanations to differ from the factual image exclusively within a predetermined region. \method{} is not only very effective in sampling such explanations, proven by new quantitative records, but also showcases the novel capabilities for explaining image classifiers enabled by RVCEs. Specifically, to properly reason about the model’s decision-making process, one must ensure that the potential confounding factors are limited to the greatest possible extent. RVCEs obtained with \method{} allow to do that in a wide range of scenarios, ranging from a fully automated approach to incorporating the user directly into the interactive explanation creation process. Our work establishes a new paradigm for explaining image classifiers in a much more principled manner, allowing the receiver to infer causally about the model’s reasoning.

\bibliography{iclr_conference}

\begin{thebibliography}{74}
\providecommand{\natexlab}[1]{#1}
\providecommand{\url}[1]{\texttt{#1}}
\expandafter\ifx\csname urlstyle\endcsname\relax
  \providecommand{\doi}[1]{doi: #1}\else
  \providecommand{\doi}{doi: \begingroup \urlstyle{rm}\Url}\fi

\bibitem[Anderson(1982)]{anderson1982reverse}
Brian~DO Anderson.
\newblock \emph{Reverse-time diffusion equation models}, volume~12.
\newblock Stochastic Processes and their Applications, Elsevier, 1982.

\bibitem[Augustin et~al.(2022)Augustin, Boreiko, Croce, and Hein]{augustin2022diffusion}
Maximilian Augustin, Valentyn Boreiko, Francesco Croce, and Matthias Hein.
\newblock Diffusion visual counterfactual explanations.
\newblock In \emph{Advances in Neural Information Processing Systems}, 2022.

\bibitem[Augustin et~al.(2024)Augustin, Neuhaus, and Hein]{augustin2024digin}
Maximilian Augustin, Yannic Neuhaus, and Matthias Hein.
\newblock Analyzing and explaining image classifiers via diffusion guidance.
\newblock In \emph{Proceedings of the IEEE/CVF Conference on Computer Vision and Pattern Recognition}, 2024.

\bibitem[Bach et~al.(2015)Bach, Binder, Montavon, Klauschen, M{\"u}ller, and Samek]{bach2015pixel}
Sebastian Bach, Alexander Binder, Gr{\'e}goire Montavon, Frederick Klauschen, Klaus-Robert M{\"u}ller, and Wojciech Samek.
\newblock On pixel-wise explanations for non-linear classifier decisions by layer-wise relevance propagation.
\newblock In \emph{PloS one}, 2015.

\bibitem[Biecek \& Samek(2024)Biecek and Samek]{pmlrv235}
Przemyslaw Biecek and Wojciech Samek.
\newblock Position: Explain to question not to justify.
\newblock In \emph{Proceedings of the 41st International Conference on Machine Learning}, volume 235, pp.\  3996--4006, 21--27 Jul 2024.

\bibitem[Boreiko et~al.(2022)Boreiko, Augustin, Croce, Berens, and Hein]{boreiko2022sparse}
Valentyn Boreiko, Maximilian Augustin, Francesco Croce, Philipp Berens, and Matthias Hein.
\newblock Sparse visual counterfactual explanations in image space.
\newblock In \emph{DAGM German Conference on Pattern Recognition}, pp.\  133--148. Springer, 2022.

\bibitem[Chang et~al.(2019)Chang, Creager, Goldenberg, and Duvenaud]{changexplaining}
Chun-Hao Chang, Elliot Creager, Anna Goldenberg, and David Duvenaud.
\newblock Explaining image classifiers by counterfactual generation.
\newblock In \emph{International Conference on Learning Representations}, 2019.

\bibitem[Chen \& He(2021)Chen and He]{chen2021exploring}
Xinlei Chen and Kaiming He.
\newblock Exploring simple siamese representation learning.
\newblock In \emph{Proceedings of the IEEE/CVF conference on computer vision and pattern recognition}, pp.\  15750--15758, 2021.

\bibitem[Chung et~al.(2022)Chung, Sim, Ryu, and Ye]{chung2022improving}
Hyungjin Chung, Byeongsu Sim, Dohoon Ryu, and Jong~Chul Ye.
\newblock Improving diffusion models for inverse problems using manifold constraints.
\newblock In \emph{Advances in Neural Information Processing Systems}, 2022.

\bibitem[Chung et~al.(2023{\natexlab{a}})Chung, Kim, Mccann, Klasky, and Ye]{chung2023diffusion}
Hyungjin Chung, Jeongsol Kim, Michael~Thompson Mccann, Marc~Louis Klasky, and Jong~Chul Ye.
\newblock Diffusion posterior sampling for general noisy inverse problems.
\newblock In \emph{The Eleventh International Conference on Learning Representations}, 2023{\natexlab{a}}.

\bibitem[Chung et~al.(2023{\natexlab{b}})Chung, Kim, and Ye]{chung2023direct}
Hyungjin Chung, Jeongsol Kim, and Jong~Chul Ye.
\newblock Direct diffusion bridge using data consistency for inverse problems.
\newblock In \emph{Thirty-seventh Conference on Neural Information Processing Systems}, 2023{\natexlab{b}}.

\bibitem[Deng et~al.(2009)Deng, Dong, Socher, Li, Li, and Fei-Fei]{deng2009imagenet}
Jia Deng, Wei Dong, Richard Socher, Li-Jia Li, Kai Li, and Li~Fei-Fei.
\newblock Imagenet: A large-scale hierarchical image database.
\newblock In \emph{2009 IEEE conference on computer vision and pattern recognition}, pp.\  248--255. IEEE, 2009.

\bibitem[Dosovitskiy et~al.(2021)Dosovitskiy, Beyer, Kolesnikov, Weissenborn, Zhai, Unterthiner, Dehghani, Minderer, Heigold, Gelly, Uszkoreit, and Houlsby]{dosovitskiy2020image}
Alexey Dosovitskiy, Lucas Beyer, Alexander Kolesnikov, Dirk Weissenborn, Xiaohua Zhai, Thomas Unterthiner, Mostafa Dehghani, Matthias Minderer, Georg Heigold, Sylvain Gelly, Jakob Uszkoreit, and Neil Houlsby.
\newblock An image is worth 16x16 words: Transformers for image recognition at scale.
\newblock In \emph{International Conference on Learning Representations}, 2021.

\bibitem[Engstrom et~al.(2019)Engstrom, Ilyas, Salman, Santurkar, and Tsipras]{robustness}
Logan Engstrom, Andrew Ilyas, Hadi Salman, Shibani Santurkar, and Dimitris Tsipras.
\newblock Robustness (python library).
\newblock In \emph{https://github.com/MadryLab/robustness}, 2019.

\bibitem[Farid et~al.(2023)Farid, Schrodi, Argus, and Brox]{farid2023latent}
Karim Farid, Simon Schrodi, Max Argus, and Thomas Brox.
\newblock Latent diffusion counterfactual explanations.
\newblock In \emph{arXiv preprint arXiv:2310.06668}, 2023.

\bibitem[Goyal et~al.(2019)Goyal, Wu, Ernst, Batra, Parikh, and Lee]{goyal2019counterfactual}
Yash Goyal, Ziyan Wu, Jan Ernst, Dhruv Batra, Devi Parikh, and Stefan Lee.
\newblock Counterfactual visual explanations.
\newblock In \emph{International Conference on Machine Learning}, pp.\  2376--2384. PMLR, 2019.

\bibitem[He et~al.(2016)He, Zhang, Ren, and Sun]{he2016deep}
Kaiming He, Xiangyu Zhang, Shaoqing Ren, and Jian Sun.
\newblock Deep residual learning for image recognition.
\newblock In \emph{Proceedings of the IEEE conference on computer vision and pattern recognition}, pp.\  770--778, 2016.

\bibitem[Heusel et~al.(2017)Heusel, Ramsauer, Unterthiner, Nessler, and Hochreiter]{heusel2017gans}
Martin Heusel, Hubert Ramsauer, Thomas Unterthiner, Bernhard Nessler, and Sepp Hochreiter.
\newblock Gans trained by a two time-scale update rule converge to a local nash equilibrium.
\newblock In \emph{Advances in neural information processing systems}, 2017.

\bibitem[Ho et~al.(2020)Ho, Jain, and Abbeel]{ho2020denoising}
Jonathan Ho, Ajay Jain, and Pieter Abbeel.
\newblock Denoising diffusion probabilistic models.
\newblock In \emph{Advances in neural information processing systems}, 2020.

\bibitem[Holzinger et~al.(2022)Holzinger, Saranti, Molnar, Biecek, and Samek]{holzinger2022explainable}
Andreas Holzinger, Anna Saranti, Christoph Molnar, Przemyslaw Biecek, and Wojciech Samek.
\newblock {Explainable AI methods-a brief overview}.
\newblock In \emph{International workshop on extending explainable AI beyond deep models and classifiers}, pp.\  13--38. Springer, 2022.

\bibitem[Jacob et~al.(2022)Jacob, Zablocki, Ben-Younes, Chen, P{\'e}rez, and Cord]{jacob2022steex}
Paul Jacob, {\'E}loi Zablocki, Hedi Ben-Younes, Micka{\"e}l Chen, Patrick P{\'e}rez, and Matthieu Cord.
\newblock Steex: steering counterfactual explanations with semantics.
\newblock In \emph{European Conference on Computer Vision}, pp.\  387--403. Springer, 2022.

\bibitem[Jeanneret et~al.(2022)Jeanneret, Simon, and Jurie]{jeanneret2022diffusion}
Guillaume Jeanneret, Lo{\"\i}c Simon, and Fr{\'e}d{\'e}ric Jurie.
\newblock Diffusion models for counterfactual explanations.
\newblock In \emph{Proceedings of the Asian Conference on Computer Vision}, pp.\  858--876, 2022.

\bibitem[Jeanneret et~al.(2023)Jeanneret, Simon, and Jurie]{jeanneret2023adversarial}
Guillaume Jeanneret, Lo{\"\i}c Simon, and Fr{\'e}d{\'e}ric Jurie.
\newblock Adversarial counterfactual visual explanations.
\newblock In \emph{Proceedings of the IEEE/CVF Conference on Computer Vision and Pattern Recognition}, pp.\  16425--16435, 2023.

\bibitem[Jeanneret et~al.(2024)Jeanneret, Simon, and Jurie]{jeanneret2024text}
Guillaume Jeanneret, Lo{\"\i}c Simon, and Fr{\'e}d{\'e}ric Jurie.
\newblock Text-to-image models for counterfactual explanations: a black-box approach.
\newblock In \emph{Proceedings of the IEEE/CVF Winter Conference on Applications of Computer Vision}, pp.\  4757--4767, 2024.

\bibitem[Kawar et~al.(2022)Kawar, Elad, Ermon, and Song]{kawar2022denoising}
Bahjat Kawar, Michael Elad, Stefano Ermon, and Jiaming Song.
\newblock Denoising diffusion restoration models.
\newblock In \emph{Advances in Neural Information Processing Systems}, 2022.

\bibitem[Khorram \& Fuxin(2022)Khorram and Fuxin]{khorram2022cycle}
Saeed Khorram and Li~Fuxin.
\newblock Cycle-consistent counterfactuals by latent transformations.
\newblock In \emph{Proceedings of the IEEE/CVF Conference on Computer Vision and Pattern Recognition}, pp.\  10203--10212, 2022.

\bibitem[Kingma(2019)]{kingmaadam}
Ba~Jimmy Kingma, Diederik~P.
\newblock Adam: A method for stochastic optimization.
\newblock In \emph{International Conference on Learning Representations}, 2019.

\bibitem[Kirillov et~al.(2023)Kirillov, Mintun, Ravi, Mao, Rolland, Gustafson, Xiao, Whitehead, Berg, Lo, et~al.]{kirillov2023segment}
Alexander Kirillov, Eric Mintun, Nikhila Ravi, Hanzi Mao, Chloe Rolland, Laura Gustafson, Tete Xiao, Spencer Whitehead, Alexander~C Berg, Wan-Yen Lo, et~al.
\newblock Segment anything.
\newblock In \emph{Proceedings of the IEEE/CVF International Conference on Computer Vision}, pp.\  4015--4026, 2023.

\bibitem[Kirsch(2011)]{kirsch2011introduction}
Andreas Kirsch.
\newblock \emph{An introduction to the mathematical theory of inverse problems}.
\newblock Applied mathematical sciences. Springer, New York, 2nd ed edition, 2011.

\bibitem[Kokhlikyan et~al.(2020)Kokhlikyan, Miglani, Martin, Wang, Alsallakh, Reynolds, Melnikov, Kliushkina, Araya, Yan, et~al.]{kokhlikyan2020captum}
Narine Kokhlikyan, Vivek Miglani, Miguel Martin, Edward Wang, Bilal Alsallakh, Jonathan Reynolds, Alexander Melnikov, Natalia Kliushkina, Carlos Araya, Siqi Yan, et~al.
\newblock Captum: A unified and generic model interpretability library for pytorch.
\newblock In \emph{arXiv preprint arXiv:2009.07896}, 2020.

\bibitem[Lang et~al.(2021)Lang, Gandelsman, Yarom, Wald, Elidan, Hassidim, Freeman, Isola, Globerson, Irani, et~al.]{lang2021explaining}
Oran Lang, Yossi Gandelsman, Michal Yarom, Yoav Wald, Gal Elidan, Avinatan Hassidim, William~T Freeman, Phillip Isola, Amir Globerson, Michal Irani, et~al.
\newblock Explaining in style: Training a gan to explain a classifier in stylespace.
\newblock In \emph{Proceedings of the IEEE/CVF International Conference on Computer Vision}, pp.\  693--702, 2021.

\bibitem[Liu et~al.(2023{\natexlab{a}})Liu, Vahdat, Huang, Theodorou, Nie, and Anandkumar]{liu20232}
Guan-Horng Liu, Arash Vahdat, De-An Huang, Evangelos Theodorou, Weili Nie, and Anima Anandkumar.
\newblock {I2SB: Image-to-Image Schr{\"o}dinger Bridge}.
\newblock In \emph{International Conference on Machine Learning}, pp.\  22042--22062. PMLR, 2023{\natexlab{a}}.

\bibitem[Liu et~al.(2023{\natexlab{b}})Liu, Zeng, Ren, Li, Zhang, Yang, Li, Yang, Su, Zhu, et~al.]{liu2024grounding}
Shilong Liu, Zhaoyang Zeng, Tianhe Ren, Feng Li, Hao Zhang, Jie Yang, Chunyuan Li, Jianwei Yang, Hang Su, Jun Zhu, et~al.
\newblock Grounding dino: Marrying dino with grounded pre-training for open-set object detection.
\newblock In \emph{arXiv preprint arXiv:2303.05499}, 2023{\natexlab{b}}.

\bibitem[Liu et~al.(2021)Liu, Lin, Cao, Hu, Wei, Zhang, Lin, and Guo]{liu2021swin}
Ze~Liu, Yutong Lin, Yue Cao, Han Hu, Yixuan Wei, Zheng Zhang, Stephen Lin, and Baining Guo.
\newblock Swin transformer: Hierarchical vision transformer using shifted windows.
\newblock In \emph{Proceedings of the IEEE/CVF international conference on computer vision}, pp.\  10012--10022, 2021.

\bibitem[Liu et~al.(2022)Liu, Mao, Wu, Feichtenhofer, Darrell, and Xie]{liu2022convnet}
Zhuang Liu, Hanzi Mao, Chao-Yuan Wu, Christoph Feichtenhofer, Trevor Darrell, and Saining Xie.
\newblock A convnet for the 2020s.
\newblock In \emph{Proceedings of the IEEE/CVF conference on computer vision and pattern recognition}, pp.\  11976--11986, 2022.

\bibitem[Lugmayr et~al.(2022)Lugmayr, Danelljan, Romero, Yu, Timofte, and Van~Gool]{lugmayr2022repaint}
Andreas Lugmayr, Martin Danelljan, Andres Romero, Fisher Yu, Radu Timofte, and Luc Van~Gool.
\newblock Repaint: Inpainting using denoising diffusion probabilistic models.
\newblock In \emph{Proceedings of the IEEE/CVF conference on computer vision and pattern recognition}, pp.\  11461--11471, 2022.

\bibitem[Lundberg \& Lee(2017)Lundberg and Lee]{10.5555/3295222.3295230}
Scott~M. Lundberg and Su-In Lee.
\newblock A unified approach to interpreting model predictions.
\newblock In \emph{Proceedings of the 31st International Conference on Neural Information Processing Systems}, 2017.

\bibitem[McCann(1997)]{MCCANN1997153}
Robert~J. McCann.
\newblock A convexity principle for interacting gases.
\newblock In \emph{Advances in Mathematics}, 1997.

\bibitem[Motzkus et~al.(2024)Motzkus, Hellert, and Schmid]{motzkus2024cola}
Franz Motzkus, Christian Hellert, and Ute Schmid.
\newblock Cola-dce--concept-guided latent diffusion counterfactual explanations.
\newblock In \emph{arXiv preprint arXiv:2406.01649}, 2024.

\bibitem[Paszke et~al.(2019)Paszke, Gross, Massa, Lerer, Bradbury, Chanan, Killeen, Lin, Gimelshein, Antiga, et~al.]{paszke2019pytorch}
Adam Paszke, Sam Gross, Francisco Massa, Adam Lerer, James Bradbury, Gregory Chanan, Trevor Killeen, Zeming Lin, Natalia Gimelshein, Luca Antiga, et~al.
\newblock Pytorch: An imperative style, high-performance deep learning library.
\newblock In \emph{Advances in neural information processing systems}, 2019.

\bibitem[Peyr\'{e} \& Cuturi(2019)Peyr\'{e} and Cuturi]{peyre2019computational}
Gabriel Peyr\'{e} and Marco Cuturi.
\newblock Computational optimal transport: With applications to data science.
\newblock \emph{Foundations and Trends in Machine Learning}, 11\penalty0 (5–6):\penalty0 355–607, 2019.

\bibitem[Preechakul et~al.(2022)Preechakul, Chatthee, Wizadwongsa, and Suwajanakorn]{preechakul2022diffusion}
Konpat Preechakul, Nattanat Chatthee, Suttisak Wizadwongsa, and Supasorn Suwajanakorn.
\newblock Diffusion autoencoders: Toward a meaningful and decodable representation.
\newblock In \emph{Proceedings of the IEEE/CVF conference on computer vision and pattern recognition}, pp.\  10619--10629, 2022.

\bibitem[Radford et~al.(2021)Radford, Kim, Hallacy, Ramesh, Goh, Agarwal, Sastry, Askell, Mishkin, Clark, et~al.]{radford2021learning}
Alec Radford, Jong~Wook Kim, Chris Hallacy, Aditya Ramesh, Gabriel Goh, Sandhini Agarwal, Girish Sastry, Amanda Askell, Pamela Mishkin, Jack Clark, et~al.
\newblock Learning transferable visual models from natural language supervision.
\newblock In \emph{International conference on machine learning}, pp.\  8748--8763. PMLR, 2021.

\bibitem[Ribeiro et~al.(2016)Ribeiro, Singh, and Guestrin]{ribeiro2016should}
Marco~Tulio Ribeiro, Sameer Singh, and Carlos Guestrin.
\newblock {"Why should I trust you?" Explaining the predictions of any classifier}.
\newblock In \emph{Proceedings of the 22nd ACM SIGKDD international conference on knowledge discovery and data mining}, pp.\  1135--1144, 2016.

\bibitem[Robbins(1992)]{robbins1992empirical}
Herbert~E Robbins.
\newblock {An empirical Bayes approach to statistics}.
\newblock In \emph{Breakthroughs in Statistics: Foundations and basic theory}, 1992.

\bibitem[Rodr{\'\i}guez et~al.(2021)Rodr{\'\i}guez, Caccia, Lacoste, Zamparo, Laradji, Charlin, and Vazquez]{rodriguez2021beyond}
Pau Rodr{\'\i}guez, Massimo Caccia, Alexandre Lacoste, Lee Zamparo, Issam Laradji, Laurent Charlin, and David Vazquez.
\newblock Beyond trivial counterfactual explanations with diverse valuable explanations.
\newblock In \emph{Proceedings of the IEEE/CVF International Conference on Computer Vision}, pp.\  1056--1065, 2021.

\bibitem[Rombach et~al.(2022)Rombach, Blattmann, Lorenz, Esser, and Ommer]{rombach2022high}
Robin Rombach, Andreas Blattmann, Dominik Lorenz, Patrick Esser, and Bj{\"o}rn Ommer.
\newblock High-resolution image synthesis with latent diffusion models.
\newblock In \emph{Proceedings of the IEEE/CVF conference on computer vision and pattern recognition}, pp.\  10684--10695, 2022.

\bibitem[Ronneberger et~al.(2015)Ronneberger, Fischer, and Brox]{ronneberger2015u}
Olaf Ronneberger, Philipp Fischer, and Thomas Brox.
\newblock U-net: Convolutional networks for biomedical image segmentation.
\newblock In \emph{Medical image computing and computer-assisted intervention--MICCAI 2015: 18th international conference, Munich, Germany, October 5-9, 2015, proceedings, part III 18}, pp.\  234--241. Springer, 2015.

\bibitem[Saharia et~al.(2022)Saharia, Chan, Chang, Lee, Ho, Salimans, Fleet, and Norouzi]{saharia2022palette}
Chitwan Saharia, William Chan, Huiwen Chang, Chris Lee, Jonathan Ho, Tim Salimans, David Fleet, and Mohammad Norouzi.
\newblock Palette: Image-to-image diffusion models.
\newblock In \emph{ACM SIGGRAPH 2022 conference proceedings}, pp.\  1--10, 2022.

\bibitem[Scarlett et~al.(2022)Scarlett, Heckel, Rodrigues, Hand, and Eldar]{scarlett2022theoretical}
Jonathan Scarlett, Reinhard Heckel, Miguel~RD Rodrigues, Paul Hand, and Yonina~C Eldar.
\newblock Theoretical perspectives on deep learning methods in inverse problems.
\newblock In \emph{IEEE journal on selected areas in information theory}, 2022.

\bibitem[Schr{\"o}dinger(1932)]{schrodinger1932theorie}
Erwin Schr{\"o}dinger.
\newblock Sur la th{\'e}orie relativiste de l'{\'e}lectron et l'interpr{\'e}tation de la m{\'e}canique quantique.
\newblock In \emph{Annales de l'institut Henri Poincar{\'e}}, volume~2, pp.\  269--310, 1932.

\bibitem[Schuhmann et~al.(2022)Schuhmann, Beaumont, Vencu, Gordon, Wightman, Cherti, Coombes, Katta, Mullis, Wortsman, et~al.]{schuhmann2022laion}
Christoph Schuhmann, Romain Beaumont, Richard Vencu, Cade Gordon, Ross Wightman, Mehdi Cherti, Theo Coombes, Aarush Katta, Clayton Mullis, Mitchell Wortsman, et~al.
\newblock Laion-5b: An open large-scale dataset for training next generation image-text models.
\newblock In \emph{Advances in Neural Information Processing Systems}, 2022.

\bibitem[Schut et~al.(2021)Schut, Key, Mc~Grath, Costabello, Sacaleanu, Gal, et~al.]{schut2021generating}
Lisa Schut, Oscar Key, Rory Mc~Grath, Luca Costabello, Bogdan Sacaleanu, Yarin Gal, et~al.
\newblock Generating interpretable counterfactual explanations by implicit minimisation of epistemic and aleatoric uncertainties.
\newblock In \emph{International Conference on Artificial Intelligence and Statistics}, pp.\  1756--1764. PMLR, 2021.

\bibitem[Selvaraju et~al.(2020)Selvaraju, Cogswell, Das, Vedantam, Parikh, and Batra]{selvaraju2020grad}
Ramprasaath~R Selvaraju, Michael Cogswell, Abhishek Das, Ramakrishna Vedantam, Devi Parikh, and Dhruv Batra.
\newblock Grad-cam: visual explanations from deep networks via gradient-based localization.
\newblock In \emph{International journal of computer vision}, 2020.

\bibitem[Shih et~al.(2021)Shih, Tien, and Karnin]{shih2021ganmex}
Sheng-Min Shih, Pin-Ju Tien, and Zohar Karnin.
\newblock Ganmex: One-vs-one attributions using gan-based model explainability.
\newblock In \emph{International Conference on Machine Learning}, pp.\  9592--9602. PMLR, 2021.

\bibitem[Shrikumar et~al.(2017)Shrikumar, Greenside, and Kundaje]{shrikumar2017learning}
Avanti Shrikumar, Peyton Greenside, and Anshul Kundaje.
\newblock Learning important features through propagating activation differences.
\newblock In \emph{International conference on machine learning}, pp.\  3145--3153. PMLR, 2017.

\bibitem[Simonyan \& Zisserman(2015)Simonyan and Zisserman]{simonyan2014very}
Karen Simonyan and Andrew Zisserman.
\newblock Very deep convolutional networks for large-scale image recognition.
\newblock In \emph{International Conference on Learning Representations}, 2015.

\bibitem[Simonyan et~al.(2014)Simonyan, Vedaldi, and Zisserman]{simonyan2013deep}
Karen Simonyan, Andrea Vedaldi, and Andrew Zisserman.
\newblock Deep inside convolutional networks: Visualising image classification models and saliency maps.
\newblock In \emph{2nd International Conference on Learning Representations, Workshop Track Proceedings}, 2014.

\bibitem[Singla et~al.(2020)Singla, Pollack, Chen, and Batmanghelich]{singlaexplanation}
Sumedha Singla, Brian Pollack, Junxiang Chen, and Kayhan Batmanghelich.
\newblock Explanation by progressive exaggeration.
\newblock In \emph{International Conference on Learning Representations}, 2020.

\bibitem[Sobieski \& Biecek(2024)Sobieski and Biecek]{sobieski2024global}
Bartlomiej Sobieski and Przemys{\l}aw Biecek.
\newblock Global counterfactual directions.
\newblock In \emph{European Conference on Computer Vision}, 2024.

\bibitem[Song et~al.(2021)Song, Sohl-Dickstein, Kingma, Kumar, Ermon, and Poole]{songscore}
Yang Song, Jascha Sohl-Dickstein, Diederik~P Kingma, Abhishek Kumar, Stefano Ermon, and Ben Poole.
\newblock Score-based generative modeling through stochastic differential equations.
\newblock In \emph{International Conference on Learning Representations}, 2021.

\bibitem[Springenberg et~al.(2015)Springenberg, Dosovitskiy, Brox, and Riedmiller]{springenberg2014striving}
Jost~Tobias Springenberg, Alexey Dosovitskiy, Thomas Brox, and Martin Riedmiller.
\newblock Striving for simplicity: The all convolutional net.
\newblock In \emph{International Conference on Learning Representations, Workshop Track Proceedings}, 2015.

\bibitem[Sundararajan et~al.(2017)Sundararajan, Taly, and Yan]{sundararajan2017axiomatic}
Mukund Sundararajan, Ankur Taly, and Qiqi Yan.
\newblock Axiomatic attribution for deep networks.
\newblock In \emph{International conference on machine learning}, pp.\  3319--3328. PMLR, 2017.

\bibitem[Szegedy et~al.(2016)Szegedy, Vanhoucke, Ioffe, Shlens, and Wojna]{szegedy2016rethinking}
Christian Szegedy, Vincent Vanhoucke, Sergey Ioffe, Jon Shlens, and Zbigniew Wojna.
\newblock Rethinking the inception architecture for computer vision.
\newblock In \emph{Proceedings of the IEEE conference on computer vision and pattern recognition}, pp.\  2818--2826, 2016.

\bibitem[Thiagarajan et~al.(2021)Thiagarajan, Narayanaswamy, Rajan, Liang, Chaudhari, and Spanias]{thiagarajan2021designing}
Jayaraman Thiagarajan, Vivek~Sivaraman Narayanaswamy, Deepta Rajan, Jia Liang, Akshay Chaudhari, and Andreas Spanias.
\newblock Designing counterfactual generators using deep model inversion.
\newblock In \emph{Advances in Neural Information Processing Systems}, 2021.

\bibitem[Tian et~al.(2022)Tian, Wu, Dai, Hu, and Jiang]{tian2022deeper}
Rui Tian, Zuxuan Wu, Qi~Dai, Han Hu, and Yu-Gang Jiang.
\newblock Deeper insights into the robustness of vits towards common corruptions.
\newblock In \emph{arXiv preprint arXiv:2204.12143}, 2022.

\bibitem[Tomaszewska \& Biecek(2024)Tomaszewska and Biecek]{tomaszewska2024position}
Paulina Tomaszewska and Przemys{\l}aw Biecek.
\newblock Position paper: Do not explain (vision models) without context.
\newblock In \emph{International Conference on Machine Learning}, 2024.

\bibitem[Vaeth et~al.(2024)Vaeth, Fruehwald, Paassen, and Gregorova]{vaeth2024gradcheck}
Philipp Vaeth, Alexander~M Fruehwald, Benjamin Paassen, and Magda Gregorova.
\newblock Gradcheck: Analyzing classifier guidance gradients for conditional diffusion sampling.
\newblock In \emph{arXiv preprint arXiv:2406.17399}, 2024.

\bibitem[Van~Looveren \& Klaise(2021)Van~Looveren and Klaise]{van2021interpretable}
Arnaud Van~Looveren and Janis Klaise.
\newblock Interpretable counterfactual explanations guided by prototypes.
\newblock In \emph{Joint European Conference on Machine Learning and Knowledge Discovery in Databases}, pp.\  650--665. Springer, 2021.

\bibitem[Weng et~al.(2024)Weng, Pegios, Feragen, Petersen, and Bigdeli]{weng2024fast}
Nina Weng, Paraskevas Pegios, Aasa Feragen, Eike Petersen, and Siavash Bigdeli.
\newblock Fast diffusion-based counterfactuals for shortcut removal and generation.
\newblock In \emph{European Conference on Computer Vision}, 2024.

\bibitem[Zeiler \& Fergus(2014)Zeiler and Fergus]{zeiler2014visualizing}
Matthew~D Zeiler and Rob Fergus.
\newblock Visualizing and understanding convolutional networks.
\newblock In \emph{Computer Vision--ECCV 2014: 13th European Conference, Zurich, Switzerland, September 6-12, 2014, Proceedings, Part I 13}, pp.\  818--833. Springer, 2014.

\bibitem[Zemni et~al.(2023)Zemni, Chen, Zablocki, Ben-Younes, P{\'e}rez, and Cord]{zemni2023octet}
Mehdi Zemni, Micka{\"e}l Chen, {\'E}loi Zablocki, H{\'e}di Ben-Younes, Patrick P{\'e}rez, and Matthieu Cord.
\newblock Octet: Object-aware counterfactual explanations.
\newblock In \emph{Proceedings of the IEEE/CVF conference on computer vision and pattern recognition}, pp.\  15062--15071, 2023.

\bibitem[Zhang et~al.(2018)Zhang, Isola, Efros, Shechtman, and Wang]{zhang2018unreasonable}
Richard Zhang, Phillip Isola, Alexei~A Efros, Eli Shechtman, and Oliver Wang.
\newblock The unreasonable effectiveness of deep features as a perceptual metric.
\newblock In \emph{Proceedings of the IEEE conference on computer vision and pattern recognition}, pp.\  586--595, 2018.

\bibitem[Zhao et~al.(2018)Zhao, Dua, and Singh]{zhao2018generating}
Zhengli Zhao, Dheeru Dua, and Sameer Singh.
\newblock Generating natural adversarial examples.
\newblock In \emph{International Conference on Learning Representations}, 2018.

\end{thebibliography}
\bibliographystyle{iclr_conference}

\newpage
\appendix
\tableofcontents

\vspace{2em}

The Appendix is structured as follows. \Cref{app:pseudocode} shows pseudocode for both \isb{} (stochastic and deterministic version) and our \method{}. \Cref{app:background_related} includes additional background knowledge connected with \isb{} and an extensive literature review regarding topics connected with our work. \Cref{app:method} shows additional figures from the method's description, considerations regarding the incorporation of the classifier's signal into \isb{} and more detailed derivation of the OT-ODE version. \Cref{app:experiments} extends our experimental evaluation with details about the setup, additional results regarding, \eg, efficiency and diversity, and concludes with details about the adaptation of other inpainting algorithms. \Cref{app:qualitative} provides qualitative examples for 7 additional classifiers, showing the versatility of \method{}, together with more RVCEs for the ResNet50 model.

\newpage

\section{Pseudocode}\label{app:pseudocode}

\begin{algorithm}[H]
    \caption{Standard I$^2$SB Generation}
    \label{alg:sample}
    \begin{algorithmic}[1]
    \STATE {\bfseries Input:} $\mathbf{x}_N \sim p_1{(\mathbf{x}_N)}$, trained $\mathbf{s}_{\boldsymbol{\psi}}{(\cdot,\cdot)}$
    \FOR{$n=N$ {\bfseries to} $1$}
    \STATE Predict $\hat{\mathbf{x}}_0(\mathbf{x}_n)$ using $\mathbf{s}_{\boldsymbol{\psi}}(\mathbf{x}_n, t_n)$
    \STATE $\mathbf{x}_{n-1} \sim p(\mathbf{x}_{n-1} \mid \hat{\mathbf{x}}_0, \mathbf{x}_n)$ according to DDPM
    \ENDFOR
    \STATE {\bfseries return} $\mathbf{x}_0$
    \end{algorithmic}
 \end{algorithm}

\begin{algorithm}[H]
    \caption{OT-ODE I$^2$SB Generation}
    \label{alg:sample-ot}
    \begin{algorithmic}[1]
    \STATE {\bfseries Input:} $\mathbf{x}_N \sim p_1{(\mathbf{x}_N)}$, trained $\mathbf{s}_{\boldsymbol{\psi}}{(\cdot,\cdot)}$
    \FOR{$n=N$ {\bfseries to} $1$}
    \STATE Predict $\hat{\mathbf{x}}_0(\mathbf{x}_t)$ using $\mathbf{s}_{\boldsymbol{\psi}}(\mathbf{x}_n, t_n)$
    \STATE $\mathbf{x}_{n-1} = \mu_{n-1} \hat{\mathbf{x}}_0 + \bar{\mu}_{n-1} \mathbf{x}_n$
    \ENDFOR
    \STATE {\bfseries return} $\mathbf{x}_0$
    \end{algorithmic}
 \end{algorithm}
 
\begin{algorithm}[H]
    \caption{RCSB}
    \label{alg:sample-rcsb}
    \begin{algorithmic}[1]
    \STATE {\bfseries Input:} Number of steps $N$, binary region mask $\mathbf{R}$,  trajectory truncation $\tau$, classifier scale $s$, input image $\mathbf{x}^*$, trained $\mathbf{s}_{\boldsymbol{\psi}}(\cdot,\cdot)$, trained classifier $f(\mathbf{y} \mid \cdot)$, target class $y$
    \STATE $\mathbf{x}_1 = (\mathbf{1}-\mathbf{R}) \odot \mathbf{x}^* + \mathbf{R} \odot \mathbf{z}$, where $\mathbf{z} \sim \mathcal{N}(\mathbf{z}; \mathbf{0}, \mathbf{I})$
    \STATE Discretize truncated timeline $0 = t_0 < t_1 < \cdots < t_N = \tau$
    \STATE $\mathbf{x}_{N} \sim q(\mathbf{x}_{N}|\mathbf{x}_0, \mathbf{x}_1)$ \hfill \# sample from analytic posterior (\cref{eq:analytic_posterior})
    \FOR{$n=N$ {\bfseries to} $1$}
    \STATE Predict $\hat{\mathbf{x}}_0(\mathbf{x}_n)$ using $\mathbf{s}_{\boldsymbol{\psi}}(\mathbf{x}_n, t_n)$
    \STATE $\mathbf{g}_n = \nabla_{\mathbf{x}_n}\log{f(y \mid {\hat{\mathbf{x}}_0})}$
    \STATE $\bar{\mathbf{g}}_n = \text{ADAM}(\mathbf{g}_n)$
    \IF{$n == N$}
        \STATE $g = \lVert \bar{\mathbf{g}}_{N} \rVert_2$ \hfill \# register norm of the first gradient
    \ENDIF 
    \STATE $\bar{\mathbf{x}}_n = \mathbf{x}_n + s \frac{\bar{\mathbf{g}}_n}{g}$
    \STATE $\mathbf{x}_{n-1} = \mu_{n-1} \hat{\mathbf{x}}_0 + \bar{\mu}_{n-1} \bar{\mathbf{x}}_n$
    \ENDFOR
    \STATE {\bfseries return} $\mathbf{x}_0$
    \end{algorithmic}
 \end{algorithm}

\begin{algorithm}[H]
    \caption{ADAM Update Rule}
    \label{alg:adam-update-rcsb}
    \begin{algorithmic}[1]
    \STATE {\bfseries Input:} Gradient at step $n$ $\mathbf{g}_n$, hyperparameters $\alpha, \boldsymbol{\epsilon}, \beta_1, \beta_2$ (set to PyTorch \citep{paszke2019pytorch} defaults)
    \STATE $\mathbf{m}_n = \beta_1 \mathbf{m}_{n-1} + (1 - \beta_1) \mathbf{g}_n$ \hfill \# update biased first moment estimate
    \STATE $\mathbf{v}_n = \beta_2 \mathbf{v}_{n-1} + (1 - \beta_2) \mathbf{g}_n^2$ \hfill \# update biased second moment estimate
    \STATE $\hat{\mathbf{m}}_n = \mathbf{m}_n / (1 - \beta_1^n)$ \hfill \# compute bias-corrected first moment
    \STATE $\hat{\mathbf{v}}_n = \mathbf{v}_n / (1 - \beta_2^n)$ \hfill \# compute bias-corrected second moment
    \STATE $\bar{\mathbf{g}}_n = \alpha \hat{\mathbf{m}}_n / (\sqrt{\hat{\mathbf{v}}_n} + \epsilon)$ \hfill \# update gradient
    \STATE {\bfseries return} $\bar{\mathbf{g}}_n$ \hfill \# return updated gradient
    \end{algorithmic}
\end{algorithm}

\begin{figure}[h]
    \centering
\includegraphics[width=0.8\linewidth]{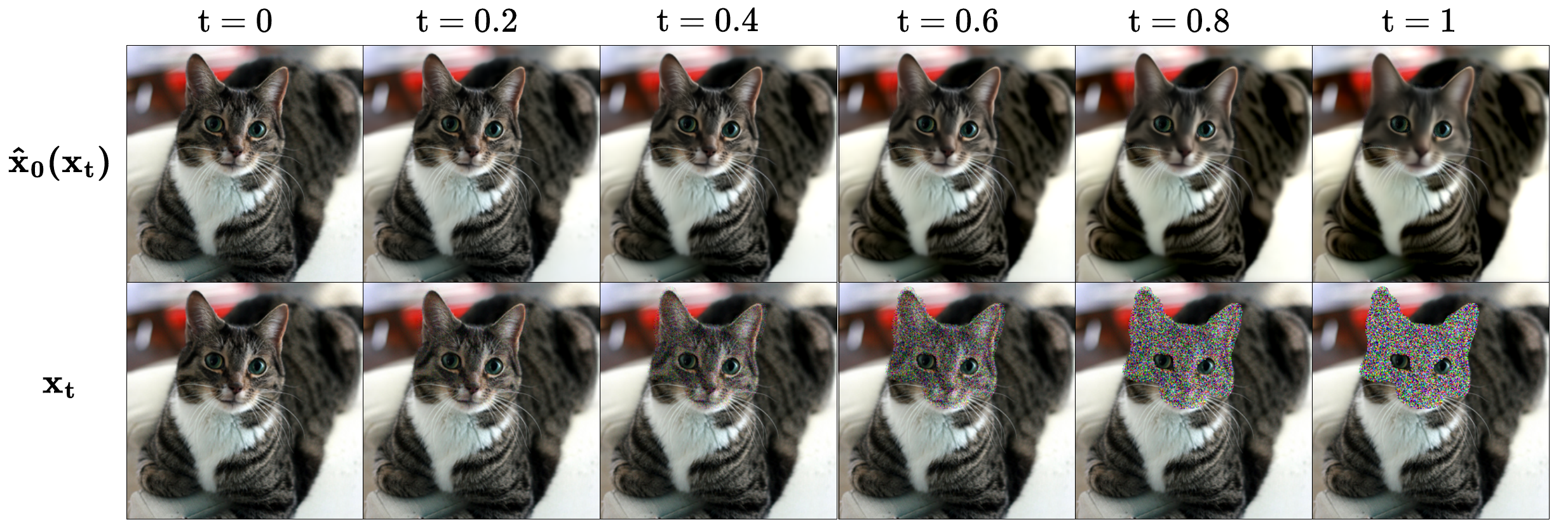}
    \caption{Visual difference between $\mathbf{x}_t$ and its corresponding Tweedie's estimate $\hat{\mathbf{x}}_0(\mathbf{x}_t)$ across different timesteps.}
    \label{fig:tweedie_difference}
\end{figure}

\begin{figure}[h]
    \centering    
    \includegraphics[width=1.0\linewidth]{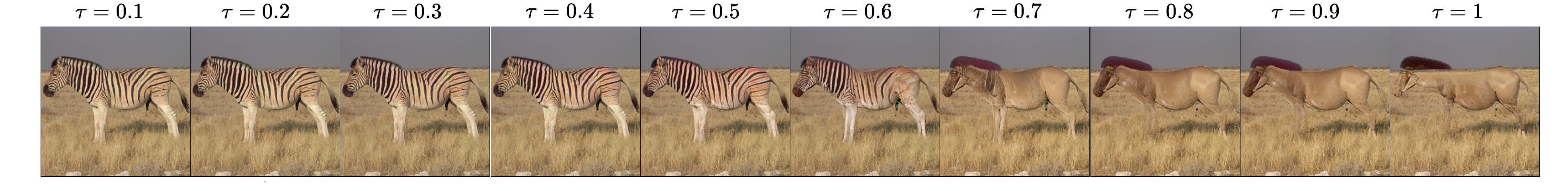}
    \caption{Influence of manipulating $\tau$ on the final RVCE obtained with the region shown in \cref{fig:3_inc_imp}.}
    \label{fig:trajectory_truncation}
\end{figure}

\newpage

\section{Extended Background \& Related Work}\label{app:background_related}

\textbf{Visual counterfactual explanations.} In recent years, increasing attention is being paid to synthesizing VCEs for image classifiers \citep{goyal2019counterfactual,schut2021generating}. These explanations aim at elucidating the model's reasoning by modifying the input image in a semantically minimal and meaningful way while flipping its prediction. Utilizing generative models for this task has historically proven to be very effective \citep{changexplaining,singlaexplanation,lang2021explaining}. Non-SGM-based methods include works like \cite{thiagarajan2021designing} which builds on the concept of deep inversion, OCTET from \citet{zemni2023octet} focusing on VCEs for complex scenes and more examples built on top of generative models \citep{rodriguez2021beyond, jacob2022steex, shih2021ganmex,zhao2018generating,van2021interpretable}.

While offering impressive results \citep{farid2023latent,jeanneret2024text,augustin2024digin,motzkus2024cola}, we argue that utilizing general foundation models like SD in the VCE generation task may cause misleading conclusions, since the explained classifiers are trained on much smaller datasets than the generative model. For example, about 1 million images from ImageNet \citep{deng2009imagenet} are used to train the classifier, while SD is trained on  5 billion images from LAION-5B \citep{schuhmann2022laion}. This discrepancy may naturally lead to SD synthesizing realistically looking variations of a given image that flip the classifier's decision but simultaneously include semantic attributes never observed by the predictive model during training. Therefore, one may question the \emph{counterfactual} nature of the explanation, as the classifier should not be expected to correctly treat attributes that it was never close to observing. Hence, in this work, we focus on generative models trained on the same data as the classifier of interest. This way, we can study the behavior of predictive models in a \emph{faithfull} manner, which is an open challenge for XAI community \cite{pmlrv235}.

\textbf{Inverse problems.} Inverse problems \citep{kirsch2011introduction} are defined as the task of recovering an unknown signal $\mathbf{x}$ based on a measurement $\mathbf{y}$ related via a measurement model $\mathbf{H}$ through $\mathbf{y} = \mathbf{H}(\mathbf{x})$, where $\mathbf{H}$ is not necessarily required to be linear or bijective. Hence, for a given measurement $\mathbf{y}$, there may exist a probability distribution over possible solutions $p( \mathbf{x} \mid \mathbf{y} = \mathbf{H}(\mathbf{x}))$. One special case of an inverse problem is image inpainting, where the missing area of an image, indicated by the mask $\mathbf{M}$, must be infilled using the available context. The measurement model is then defined as $\mathbf{H}(\mathbf{x}) = \mathbf{M} \odot \mathbf{x}$, where $\odot$ denotes an element-wise product and $\mathbf{M}$ is a binary mask.

In recent years, deep learning methods have proven to be very effective at solving various kinds of inverse problems \citep{scarlett2022theoretical}. Recently, utilizing generative methods, especially SGMs, established itself as the new SOTA approach in the image domain. One way of adapting SGMs to inverse problems is through conditional generation, where the conditional score can be derived with the measurement model. Many additional techniques, such as data consistency \citep{chung2023direct}, manifold constraint \citep{chung2022improving,chung2023diffusion} and others \citep{kawar2022denoising}, are further utilized to improve this adaptation. 

\textbf{Image-to-Image Schrödinger Bridges.} A much harder but possibly also much more effective approach is to learn \emph{direct} mappings between the distribution of signals $\mathbf{x} \sim p_0$ and measurements $\mathbf{y} \sim p_1$ instead of adapting pretrained models. In this line of research, \citet{liu20232} propose to learn such mappings by constructing a tractable subclass of Schrödinger bridges (SBs, \citet{schrodinger1932theorie}), termed Image-to-Image Schrödinger Bridges (\isbs{}). The SB is an entropy-regularized optimal transport model, which, resembling the framework of SGMs, considers the following forward and backward SDEs:
\begin{subequations}
\begin{align}
    \diff \mathbf{x}_t &= (\mathbf{F}_t(\mathbf{\mathbf{x}_t}) + \beta_t \nonlscoret) \diff t + \sqrt{\beta_t} \diff \mathbf{w}, \label{eq:sb_forward_sde} \\
    \diff \mathbf{x}_t &= (\mathbf{F}_t(\mathbf{\mathbf{x}_t}) - \beta_t \nonlhscoret) \diff t + \sqrt{\beta_t} \diff \mathbf{\Bar{w}},\label{eq:sb_reverse_sde}
\end{align}
\end{subequations}

Similarly to SGMs, the marginal densities of \cref{eq:sb_forward_sde,eq:sb_reverse_sde} are equivalent. The functions $\boldpsi,\boldhatpsi \in C^{2,1}(\mathbb{R}^d, [0, 1])$ represent time-varying energy potentials and are additionally constrained to solve the following partial differential equations
\begin{subequations}
\begin{align}
   & \begin{cases}
  \frac{\boldpsi(\mathbf{x}_t, t)}{\partial t} = - \nabla \boldpsi^\top \mathbf{F} - \frac{1}{2} \beta \Delta \boldpsi \\
  \frac{\boldhatpsi (\mathbf{x}_t, t)}{\partial t} = - \nabla \cdot (\boldhatpsi \mathbf{F}) + \frac{1}{2} \beta \Delta \boldhatpsi
  \end{cases} \label{eq:sb_pdes}
  \\
   \text{s.t. }
  \boldpsi(\mathbf{x}_0,0) &  \boldhatpsi(\mathbf{x},0){=}p_0(\mathbf{x}),
      \boldpsi(\mathbf{x},1) \boldhatpsi(\mathbf{x},1){=}p_1(\mathbf{x}) \label{eq:sb_coupling}
\end{align}
\label{eq:sb_pdes_coupling}
\end{subequations}
In general, numerically solving \cref{eq:sb_forward_sde,eq:sb_reverse_sde} is much more difficult compared to SGMs due to nonlinear terms $\boldpsi, \boldhatpsi$ being coupled via \cref{eq:sb_coupling}. However, with \textbf{Theorem 3.1}, \citet{liu20232} show an important connection between the frameworks of SBs and SGMs. We repeat it here explicitly for later reference.
\begin{theorem}[Reformulating SB drifts as score functions \citep{liu20232}]
If $\boldhatpsi,\boldpsi$ fulfill the constraints given by \cref{eq:sb_pdes_coupling}, then $\nonlhscoret,\nonlscoret$ are the score functions of the following linear SDEs, respectively
\begin{align}
      \diff \mathbf{x}_t = \mathbf{F}_t(\mathbf{\mathbf{x}_t}) \diff t + \sqrt{\beta_t} \diff \mathbf{w}, \quad \mathbf{x}_0 \sim \boldhatpsi(\cdot, 0), \label{eq:i2sb_forward_sde}
      \\
      \diff \mathbf{x}_t = \mathbf{F}_t(\mathbf{\mathbf{x}_t}) \diff t + \sqrt{\beta_t} \diff \mathbf{\Bar{w}}, \quad \mathbf{x}_1 \sim \boldpsi(\cdot, 1). \label{eq:i2sb_reverse_sde}
\end{align}
\label{th:liu2023}
\end{theorem}
Crucially, \cref{th:liu2023} states that, while $\boldhatpsi,\boldpsi$ are not in general assumed to be valid probability distributions, it is true that $\nabla \log{\boldhatpsi}=\nabla \log{p^{\ref{eq:i2sb_forward_sde}}}$ and $\nabla \log{\boldpsi}=\nabla \log{p^{\ref{eq:i2sb_reverse_sde}}}$ for $p^{\ref{eq:i2sb_forward_sde}}, p^{\ref{eq:i2sb_reverse_sde}}$ representing the densities of the respective SDEs. Following this theoretical result, \citet{liu20232} also show a principled approach for approximating $\nonlhscoret$ with a neural network $\mathbf{s}_{\boldsymbol{\psi}}$. In essence, these results allow to train \emph{direct} inverse problem solvers with the use of paired data, where $p_0$ represents a clean data distribution and $p_1$ the distribution of its corrupted measurements.  Additionally, \citet{liu20232} show how \isb{} connects with flow-based optimal transport (OT) \citep{peyre2019computational,MCCANN1997153}, where assuming that $\beta_t \rightarrow 0$ leads to an ordinary differential equation (ODE) $\diff \mathbf{x}_t = \mathbf{v}_t(\mathbf{x}_t \mid \mathbf{x}_0 ) \diff t$ that provides a deterministic mapping with the use of $\mathbf{s}_{\boldsymbol{\psi}}$ estimate. In practive, this is achieved by eliminating the noise from the intermediate sampling steps (see \cref{alg:sample-ot}).

\textbf{Visual attribution methods.} The very first works in the current era of Explainable Artificial Intelligence (XAI) were concerned with providing explanations of the model's decision through visual heatmaps, which highlighted pixels considered important to the its prediction. One of the first approaches by \citet{simonyan2013deep} proposed simple backpropagation of the model's output w.r.t. the input, indicating the direction of its greatest ascent in the pixel space, often termed as Saliency. More sophisticated approaches emerged in the following years, where techniques like Layer-wise Relevance Propagation (LRP, \citet{bach2015pixel}), Integrated Gradients (IG, \citet{sundararajan2017axiomatic}), DeepLift and Input $\times$ Gradient \citep{shrikumar2017learning}, Guided Backpropagation (GuidedBackprop, \citet{springenberg2014striving}), GradCAM \citep{selvaraju2020grad}, Deconvolution \citep{zeiler2014visualizing} and others that utilize the gradient of the neural network promised to indicate more semantically meaningful concepts in a less 'noisy' manner. Concurrent line of research about perturbation-based methods assumed a more general black-box scenario, where explanations could be provided for a broader class of models. There, methods like Occlusion \citep{zeiler2014visualizing}, Local Interpretable Model-agnostic Explanations (LIME, \citet{ribeiro2016should}), SHapley Additive exPlanations (SHAP, \citet{10.5555/3295222.3295230}) and its variations quickly advanced the state-of-the-art. In this work, we utilize their unified implementations provided by the Captum package \citep{kokhlikyan2020captum}.

\section{Extended Method}\label{app:method}

\subsection{Additional figures}

We provide illustrative examples for the visual differences between $\mathbf{x}_t$ and the Tweedie's estimate $\hat{\mathbf{x}}_0(\mathbf{x}_t)$ in \cref{fig:tweedie_difference}. For the effect of manipulating the $\tau$ hyperparameter, see \cref{fig:trajectory_truncation}

\subsection{Incorporating the classifier's signal}

In the following, we explicitly define the DDPM \citep{ho2020denoising} sampler mentioned in \cref{alg:sample} and elaborate on the exact way of incorporating the classifier's gradients into the generation process of \isb{}.

Denote by $\{ t_i\}_{i\in\{0, ..., N\}}$ the discrete sequence of timesteps of length N such that $0 = t_0 < t_1 < \cdots < t_N = 1$. By $\sigma^2_{n} = \int_{0}^{t_n}{\beta_\tau d\tau}$ and $\bar{\sigma}^2_{n} = \int_{t_n}^{1}{\beta_\tau d\tau}$, we denote the variances accumulated from each side. Additionally, let $\alpha^2_{n-1} = \int_{t_{n-1}}^{t_n}{\beta_\tau d\tau}$ be the variance accumulated between two consecutive timesteps. For ease of notation, we define $\mu_{n-1}$ and $\bar{\mu}_{n-1}$ as 
\begin{align}
    \mu_{n-1} &= \frac{\alpha^2_{n-1}}{\alpha^2_{n-1} + \sigma^2_{n-1}}, \\
    \bar{\mu}_{n-1} &= \frac{\sigma^2_{n-1}}{\alpha^2_{n-1} + \sigma^2_{n-1}}.
\end{align}
With that, we can define the DDPM posterior sampler  as
\begin{align}
    \mathbf{x}_{n-1} &\sim p(\mathbf{x}_{n-1} | \hat{\mathbf{x}}_0, \mathbf{x}_n), \\
    \mathbf{x}_{n-1} &\sim \mathcal{N}\left(\mu_{n-1} \hat{\mathbf{x}}_0 + \bar{\mu}_{n-1} \mathbf{x}_n, \frac{\alpha^2_{n-1}\sigma^2_{n-1}}{\alpha^2_{n-1} + \sigma^2_{n-1}} I \right),
\end{align}
where $\hat{\mathbf{x}}_0 = \hat{\mathbf{x}}_0(\mathbf{x}_n)$ denotes the Tweedie's estimate obtained with $\mathbf{s}_{\boldsymbol{\psi}}$, \ie, a trained \isb{}.

When using the OT-ODE version of \isb{}, we replace sampling from the posterior with a deterministic version by following the mean, which yields the update rule
\begin{equation}
    \mathbf{x}_{n-1} = \mu_{n-1} \hat{\mathbf{x}}_0 + \bar{\mu}_{n-1} \mathbf{x}_n.
\end{equation}
By converting the Tweedie's estimate to the conditional score using \cref{eqn:tweedie} and applying Bayes' Theorem, we are left with 
\begin{equation}
\begin{split}
    \mathbf{x}_{n-1} &= \mu_{n-1} \left( \mathbf{x}_n + \sigma^2_n \scorenwithcond \right) + \bar{\mu}_{n-1} \mathbf{x}_n \\
     &= \mu_{n-1} \left( \mathbf{x}_n + \sigma^2_n \scoren + \sigma^2_n \scorencond \right) + \bar{\mu}_{n-1} \mathbf{x}_n,
    \label{eqn:method_update_rule}
\end{split}
\end{equation}
where $\scoren$ can be approximated by a standard I$^2$SB network trained on the task of inpainting. By manipulating \cref{eqn:method_update_rule} further, one can arrive at the following update rule
\begin{equation}
\begin{split}
    \mathbf{x}_{n-1} & = \mu_{n-1} \left( \mathbf{x}_n + \sigma^2_n \scoren \right) + \mu_{n-1} \sigma^2_n \scorencond  + \bar{\mu}_{n-1} \mathbf{x}_n \\ 
    & = \mu_{n-1} \left( \mathbf{x}_n + \sigma^2_n \scoren \right) + \bar{\mu}_{n-1} \left( \frac{\mu_{n-1} \sigma^2_n}{\bar{\mu}_{n-1}} \scorencond  +  \mathbf{x}_n \right) \\
    & =  \mu_{n-1} \left( \mathbf{x}_n + \sigma^2_n \scoren \right) + \bar{\mu}_{n-1} \left( c_n \scorencond  + \mathbf{x}_n \right).
    \label{eqn:method_update_rule_better}
\end{split}
\end{equation}
Here, we explicitly define the time-dependent coefficient $c_n$. While plugging $\nabla_{\mathbf{x}_n} \log{f(\mathbf{y} \mid \mathbf{x}_n)}$ in place of $\scorencond$ in \cref{eqn:method_update_rule} is the most intuitive, we empirically verified that replacing $c_n \scorencond$ with $\nabla_{\mathbf{x}_n} \log{f(\mathbf{y} \mid \mathbf{x}_n)}$ leads to more semantically meaningful results. Practically, this can be explained by $\mu_{n-1}$ achieving its highest values at the end of the generation process, effectively incorporating the classifier's signal to the highest extent in the final steps of the generation. Since we are interested in influencing the generative trajectory with the classifier $f$ along the entire process (and possibly decreasing its influence to the greatest possible extent in the final timesteps to avoid adversarial changes), it seems intuitive that incorporating $f$ into \cref{eqn:method_update_rule_better} allows for obtaining more meaningful RVCEs. This is due to $\bar{\mu}_{n} = 1 - \mu_{n}$, meaning that the classifier's signal is amplified in the beginning of the generation and decreased in the final steps. This intervention also explains the effectiveness of the introduced improvements, as they break the independence of the classifier's signal between consecutive steps, practically incorporating the time-dependent coefficient $c_n$ into the gradient alignment.

\subsection{Analytic posterior and OT-ODE}



Following the original work of \citet{liu20232} (\isb{}), the analytic posterior from the forward stochastic process, which governs the mapping between a given boundary pair $(\mathbf{x}_0,\mathbf{x}_1)$, is defined as 
\begin{equation}
q(\mathbf{x}_t | \mathbf{x}_0, \mathbf{x}_1) = \mathcal{N}\Big(\boldsymbol{\mu}(\mathbf{x_0}, \mathbf{x_1}, t) = \mathbf{x}_0 +  t (\mathbf{x}_1 - \mathbf{x}_0), \mathbf{\Sigma}_t = \alpha t(1-t) \mathbf{I}\Big), 
\label{eq:analytic_posterior}
\end{equation}
where by default $\alpha=1$. To arrive at the OT-ODE version of \isb{}, one must use $\alpha \rightarrow 0$, effectively reducing $q$ to a Dirac delta distribution centered at $\boldsymbol{\mu}(\mathbf{x_0}, \mathbf{x_1}, t)$.

\section{Extended Experiments}\label{app:experiments}

We follow the evaluation protocol from previous works for VCEs on Imagenet, which, for a given task, uses all images from the training subset correctly predicted by the evaluated model. For ResNet50, this results in around 2000 images per task. We extract the results of other methods from the work of \citet{farid2023latent}, except the DVCE method \citep{augustin2022diffusion}, which evalutes with a protocol that we were not able to fully reproduce. Hence, to ensure fair comparison, we adapted the implementation of DVCE to our evaluation. Specifically, we utilize the multiple-norm robust ResNet50 from the work of \citet{boreiko2022sparse}, which the authors of DVCE propose as default, to achieve VCEs for the ResNet50 model. In terms of hyperparameters, we fine-tuned them with grid search on a subset of Zebra--Sorrel task and used $s=18.0$ as the guidance scale for the non-robust ResNet50, since it performed the best.

For \isb{}, we utilize the original checkpoint from \citet{liu20232} trained on $20-30\%$ freeform masks from \citet{saharia2022palette}. While the checkpoint trained on the $10-20\%$ variant is also available and verified to work within our framework, we discovered that the former generalizes well to smaller area values. Hence, for the sake of simplicity, we utilize the $20-30\%$ version only. By default, we use NFE=100, which we explored the most, but lower NFE regimes provided promising initial results. For the automated region extraction, we use IG by default, but evaluate 10 other attribution methods in \cref{app:other_attribution_methods}.

\subsection{Details of individual experiments}

\cref{fig:3_inc_imp}: Each improvement, together with the naive approach, is evaluated on the zebra-sorrel task with around 2000 images from ImageNet training set (following the protocol from the main experimental evaluation). Each image is initially predicted as either zebra or sorrel by the ResNet50 \citep{he2016deep} model and the decision must be flipped to the opposite class. FID is computed between the obtained explanations and original images. The hyperparameter values used for all improvements are: $a=0.3$, $s=1.0$ (except ADAM stabilization with $s=1e-2$), $c=16$, $\tau=1.0$ (except trajectory truncation, where $\tau=0.6$).

\cref{tab:additional_results}: \textbf{A}: For all tasks, we use $s=1.5$ and $\tau=0.4$ (to better preserve the original content). As we cannot control the area of masks provided by LangSAM, hyperparameters $a$ and $c$ are not applicable in this scenario. Images with masks covering area greater than $40\%$ of the total image are discarded from the evaluation to ensure that we only use meaningful RVCEs. \textbf{B}: Across all tasks, the $10\%-20\%$ experiment uses configuration $B$ from \cref{tab:main_results}, while the $20\%-30\%$ experiment uses configuration $C$. Hyperparameters $a$ and $c$ are not applicable, since masks are provided automatically from the mentioned dataset. \textbf{C}: Each inpainting algorithm is given a 24 A100 GPU hours time budget, resulting in around 2000 images for DDRM, 800 images for MCG and 400 images for RePaint on each task. Details of their adaptations are provided separately in \cref{app:adaptation}. 

\subsection{Metrics description}

In the following, we provide detailed description of each metric used in the quantitative evaluation.

\textbf{FID and sFID (realism).} Following works on image synthesis, measuring the realism of the obtained explanations at a distribution level is often done with FID and sFID \citep{heusel2017gans}. Specifically, FID compares a set of real (\emph{r}) and generated (\emph{g}, in this case, the explanations) images by first extracting their corresponding features from the InceptionV3 network \citep{szegedy2016rethinking} and then computing
\begin{equation}
    \text{FID} = ||\boldsymbol{\mu}_r - \boldsymbol{\mu}_g||^2 + \text{Tr}\left( \Sigma_r + \Sigma_g - 2\left(\boldsymbol{\Sigma}_r \boldsymbol{\Sigma}_g \right)^{1/2} \right),
\end{equation}
where $\boldsymbol{\mu}_r, \boldsymbol{\mu}_g$ are the mean vectors and $\boldsymbol{\Sigma}_r, \boldsymbol{\Sigma}_g$ are the covariance matrices of the respective distributions in the feature space. As comparing original images with their edited versions (\eg, explanations) may bias the metric with original pixels mostly unchanged, artificially boosting the realism evaluation, sFID first divides the sets into folds and averages FID over the independent counterparts.

\textbf{S$^3$ (representation similarity).}

Explanations should also resemble original images from a representation respective. Here, following the work of \citet{jeanneret2023adversarial}, we compute average SimSiam Similarity (S$^3$) over a set of original images and the resulting counterfactuals. Specifically, S$^3$ utilizes a SimSiam network \citep{chen2021exploring}, which encodes both the factual and counterfactual images into their respective representations $\mathbf{r}_{f}$, $\mathbf{r}_{cf}$ and computes the cosine similarity as
\begin{equation}
    \text{S}^3 = \frac{\mathbf{r}_{f} \cdot \mathbf{r}_{cf}}{||\mathbf{r}_{f}||_2 \cdot ||\mathbf{r}_{cf}||_2}.
\end{equation}

\textbf{COUT (sparsity).} In the context of VCEs, sparsity is understood as perturbing a minimal number of pixels to flip the model's decision. To quantify this criterion, the COUnterfactual Transition (COUT) metric computes
\begin{equation}
    \begin{split}
        \text{COUT} =& \text{AUPC}_y - \text{AUPC}_{y^*}, \\
        \text{AUPC}_k =& \frac{1}{M} \sum_{m=0}^{M-1} \frac{1}{2}(f(k \mid \mathbf{x}^{(m)}) - f(k \mid \mathbf{x}^{(m+1)}))
    \end{split}
\end{equation}
where $\mathbf{x}^{(0)}$ is the factual image, $\mathbf{x}^{(M)}$ the resulting VCE, and $y^*, y$ are the class labels predicted by $f$ for $\mathbf{x}^{(0)}, \mathbf{x}^{(M)}$ respectively. In practice, COUT measures how fast the classifier's decision changes when interpolating between the original and the explanation, but the interpolation is defined as inserting pixels to the original image according to the extent (absolute value) of change observed in the VCE through $M$ steps. COUT is typically reported as an average over a set of samples.

\textbf{Flip Rate (efficiency).} A major criterion for a VCE method is its efficiency, understood as the ability to effectively flip the model's decision. For a set of triplets $\{ \mathbf{x}^*_{i}, \mathbf{x}_{i}, y_{i} \}_{i=1}^I$, where $\mathbf{x}^*_{i}$ is the original image and $\mathbf{x}_{i}$ is the resulting VCE targeted to flip $f$'s decision to $y_{i}$, Flip Rate (FR) is defined as the fraction of cases which correctly flipped the decision to the target class, \ie,
\begin{equation}
    \text{FR} = \frac{1}{I} \sum_{i=1}^I \mathbf{1}(\argmax_{y'} f(y' \mid \mathbf{x}_{i}) = y_{i}).
\end{equation}

\subsection{Adaptation of other inpainting algorithms}\label{app:adaptation}

In this subsection, we describe the adaptation of each inpainting algorithm from our ablation study. For each method, we follow the notation from its corresponding original work to make the description easier to follow.

\subsubsection{Manifold Constrained Gradient (MCG, \citet{chung2022improving})}

MCG iteratively denoises (inpaints) the missing parts with the following two-step update (Equations 14 and 15, \citet{chung2022improving}):

\begin{align}
    \mathbf{x}_{t-1}' &= \mathbf{f}(\mathbf{x_i}, \mathbf{s_\theta}) - \frac{\partial}{\partial \mathbf{x_t}}\lVert \textbf{W}(\textbf{y} - \textbf{H}\mathbf{\hat{x}_0}(\mathbf{x_t}) \rVert_2^2 + g(\mathbf{x_t})\mathbf{z}, \quad  \mathbf{z} \sim \mathcal{N}(0, \textbf{I})
     \label{eq:mcg_manifold_constraint_update}
    \\
    \mathbf{x_{i-t}} &= \textbf{A}\mathbf{x'}_{t-1} + \textbf{b}
    \label{eq:mcg_data_consistency_step}
\end{align}

where \cref{eq:mcg_manifold_constraint_update} is a manifold constraint update and \cref{eq:mcg_data_consistency_step} is a data consistency step. As described by the authors, both steps are crucial to ensure that the gradient of the measurement term stays on the manifold and to deal with the potential deviation from the measurement consistency.

Since $\mathbf{f}(\mathbf{x}_{t}, \mathbf{s}_{\boldsymbol{\theta}})$ implicitly predicts the mean $\mu_t$ and variance $\sigma_t$ at each step $t$, related to the underlying SDE dynamics, we apply our guidance scheme by modifying the original

\begin{equation}
\mathbf{f}(\mathbf{x_i}, \mathbf{s_\theta}) = \mu_t + \sigma_t \mathbf{z}, \quad  \mathbf{z} \sim \mathcal{N}(0, \textbf{I})
\end{equation}

by adding properly scaled (according to the relationship between the likelihood score and mean) conditioning:

\begin{equation}
\mathbf{f'}(\mathbf{x_i}, \mathbf{s_\theta}) = \mu_t + \sigma_t \mathbf{z} + s \cdot \sigma_t^2\frac{\bar{\mathbf{g}}_n}{g}, \quad  \mathbf{z} \sim \mathcal{N}(0, \textbf{I})
\end{equation}
where $\mathbf{g}_n$ and $g$ are obtained as described in \cref{alg:sample-rcsb}.

\subsubsection{Denosing Diffusion Restoration Models (DDRM, \citet{kawar2022denoising})}

DDRM considers the SVD decomposition of the measurement model matrix $\mathbf{H}$ in the linear noisy inverse problem
\begin{equation}
    \mathbf{y} = \mathbf{H}\mathbf{x} + \sigma_{\boldsymbol{y}} \mathbf{z}, \mathbf{z} \sim \mathcal{N}(0, \textbf{I}),
\end{equation}
where $\sigma_{\boldsymbol{y}}$ is the standard divination of the measurement noise. 
For the task of inpaiting, the $\mathbf{H}$ matrix is a diagonal matrix with either $0$ or $1$ on the diagonal indicating available and missing pixels. Hence, its SVD decomposition simplifies to using identity matrices in place of $\mathbf{U}$ and $\mathbf{V}$.

The main contribution of DDRM is that it provides a way to include the information from that decomposition and observation $\mathbf{y}$ into the generative process, which the authors summarize in Equations 7 and 8 in the original work. The method uses a trained denoising network to obtain a prediction of $\mathbf{x}_0$ at timestep $t$, denoted as $\color{violet}{\mathbf{x}_{\theta, t}}$. In order to include the additional information from the classifier into the DDRM framework, we modify that prediction to include the model's gradients by replacing the update rule
\begin{equation}
    \color{violet} \mathbf{\bar x}_{\theta, t} \color{black} =  V^T \color{violet} \mathbf{x}_{\theta, t}
\end{equation}
with
\begin{equation}
    \color{violet} \mathbf{\bar x}_{\theta, t} \color{black} = V^T (\color{violet} \mathbf{x}_{\theta, t} \color{black} + s \frac{\bar{g}}{g}),
\end{equation}
where $\bar{g}$ and $g$ are obtained as described in \cref{alg:sample-rcsb}.

\subsubsection{RePaint \citep{lugmayr2022repaint}}

RePaint performs the task of inpainting by modifying the standard denoising process, where, at each timestep $t$, the network's input is composed of noised pixels known from the original input sampled directly from $q$ and the unknown noisy pixels predicted by the network in the previous timestep. Additionally, to harmonize the two parts of the image, RePaint samples $x_t$ directly from $q(x_t| x_{t-1})$ and repeats the forward procedure. By default, this resampling scheme is repeated $20$ times for each of the standard diffusion steps.

In order to incorporate the information from the classifier, we modify the unconditional mean of the posterior $p_\theta$ in the denosing step to conditional one, effectively replacing the mean predictor
\begin{equation}
  \boldsymbol{\varepsilon}_{\boldsymbol{\theta}}(X_t, t)  
\end{equation}
shown in the $7$-th step of Algorithm $1$ from the original work \citep{lugmayr2022repaint} with
\begin{equation}
  \boldsymbol{\varepsilon}_{\boldsymbol{\theta}}(X_t, t) + s \cdot \sigma^2_t \frac{\bar{g}}{g}
\end{equation}
where $\bar{g}$ and $g$ are obtained as described in \cref{alg:sample-rcsb}.

\subsection{Schedulers for guidance scale}\label{app:schedulers}

\begin{figure}[h]
    \centering
    \begin{subfigure}{0.27\textwidth}
        \centering
        \includegraphics[width=\linewidth]{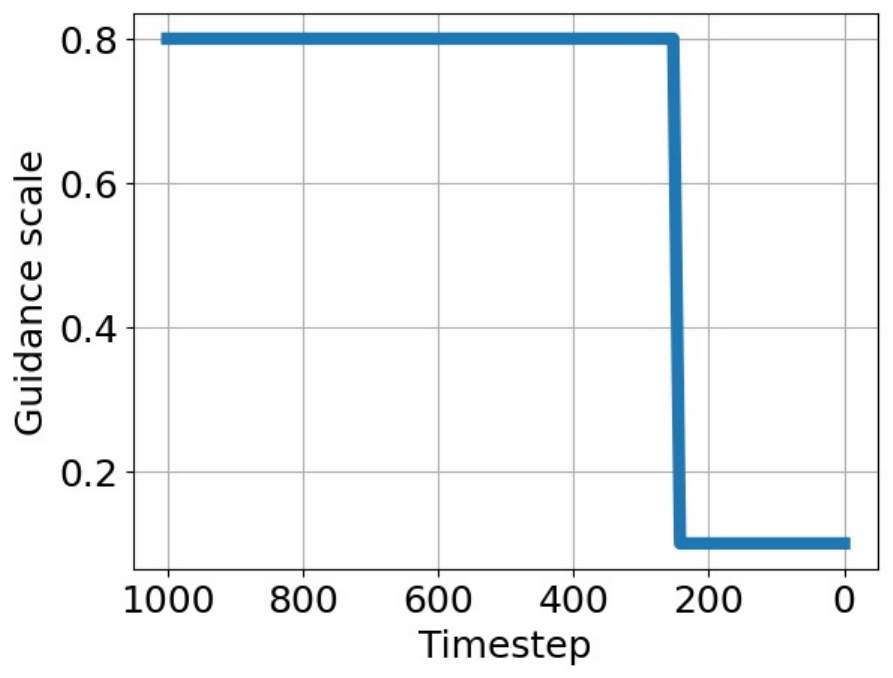} 
        \caption{Interval scheduler}
        \label{fig:schedulers1}
    \end{subfigure}
    \begin{subfigure}{0.27\textwidth}
        \centering
        \includegraphics[width=\linewidth]{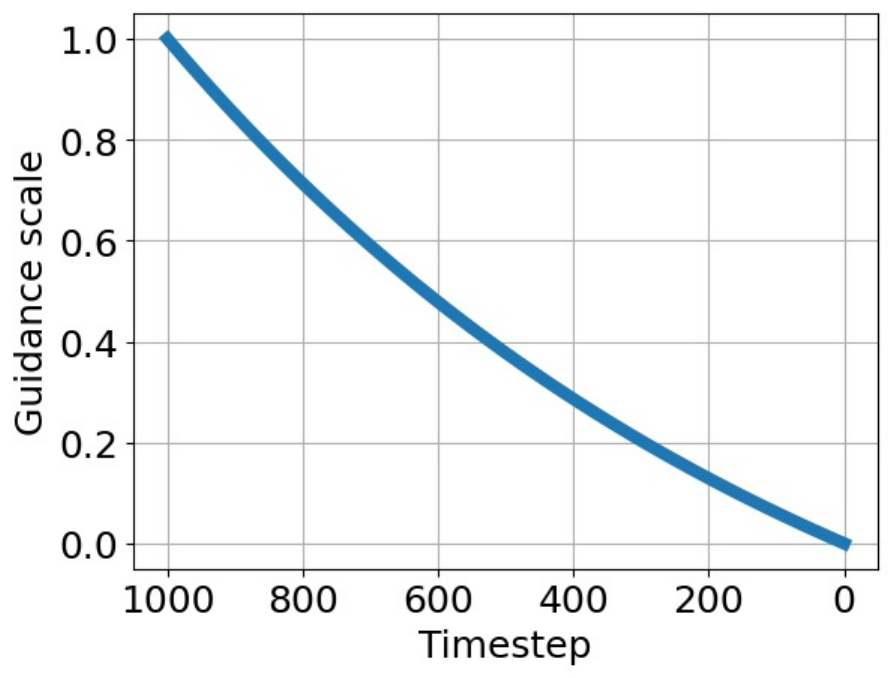}
        \caption{Exponential scheduler}
        \label{fig:schedulers2}
    \end{subfigure}
    \begin{subfigure}{0.27\textwidth}
        \centering
        \includegraphics[width=\linewidth]{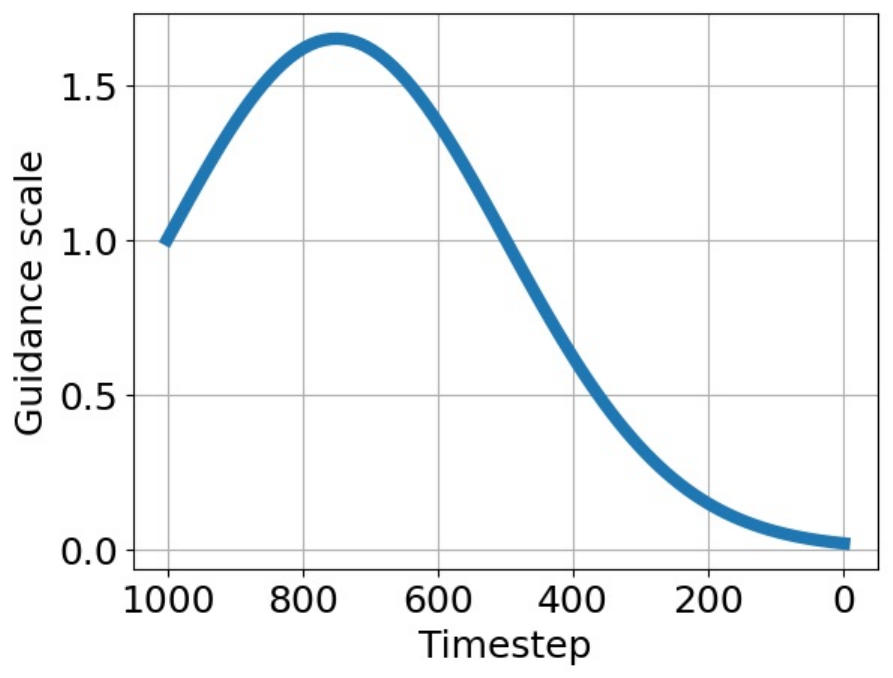}
        \caption{Gaussian scheduler}
        \label{fig:schedulers3}
    \end{subfigure}
    \caption{Visualization of more complex schedulers used throughout the development of our method.}
    \label{fig:schedulers}
\end{figure}

\Cref{fig:schedulers} visualizes example schedulers used throughout the development of our method. Our adaptive normalization technique empirically outperformed all tested schedulers.

\subsection{Quantitative evaluation of other attribution methods}\label{app:other_attribution_methods}

To pick a default attribution method for \method{}, we evaluated it on the Zebra--Sorrel using the \method{}$^B$ hyperparameter configuration for $11$ different attribution methods shown in \cref{table:attr_grid_ablation}. Based on these results, we chose Integrated Gradients \citep{sundararajan2017axiomatic} as the default, since it provides the most balanced performance. 

\begin{table}[ht]
\centering
    \scriptsize
    \begin{tabular}{c|*{5}{c}}
    \toprule
    \multicolumn{6}{c}{\textbf{Zebra -- Sorrel}} \\
    \midrule
    Attribution method & FID & sFID & S$^3$ & COUT & FR \\
    \midrule
    LRP & $\mathbf{7.5}$ & $\mathbf{15.5}$ & $\mathbf{0.87}$ & $0.62$ & $93.6$ \\
    InputXGradient & $\underline{9.0}$ & $\underline{16.8}$ & $\underline{0.87}$ & $0.73$ & $\underline{97.8}$ \\
    DeepLift & $9.2$ & $17.0$ & $0.87$ & $0.73$ & $\mathbf{97.9}$ \\
    Integrated Gradients & $9.5$ & $17.4$ & $0.86$ & $0.72$ & $97.4$ \\
    GradientShap & $10.5$ & $18.5$ & $0.87$ & $\underline{0.74}$ & $97.4$ \\
    LIME & $12.9$ & $20.7$ & $0.85$ & $0.55$ & $88.4$ \\
    GuidedBackprop & $13.8$ & $21.49$ & $0.86$ & $0.72$ & $96.5$ \\
    Occlusion & $13.9$ & $21.7$ & $0.86$ & $0.50$ & $86.0$ \\
    GradCAM & $14.1$ & $22.15$ & $0.85$ & $0.52$ & $87.1$ \\
    GuidedGradCAM & $15.1$ & $22.5$ & $0.86$ & $0.71$ & $96.1$ \\
    Saliency & $15.2$ & $23.0$ & $0.86$ & $\bold{0.75}$ & $98.4$ \\
    \bottomrule
    \end{tabular}
    \caption{Quantitative evaluation of 11 attribution methods (described in \cref{app:background_related}) on the Zebra--Sorrel task following our evaluation protocol.}
    \label{table:attr_grid_ablation}
\end{table}

\subsection{Diversity assessment}

\method{} utilizes the OT-ODE version of the \isb{}, which provides a deterministic mapping between the noisy image and the resulting RVCE. The source of randomness comes from the Gaussian noise inserted into the image in the place of missing pixels at the beginning of the inpainting process. In order to examine the diversity of the generated RVCEs, we followed the evaluation procedure from the work of \citet{jeanneret2023adversarial}. In essence, we compute the mean pair-wise LPIPS metric between two runs with different seeds (used in generation of the Gaussian noise) for our three main configurations of hyperparameters \method{}$^A$, \method{}$^B$ and \method{}$^C$. For each run, $256$ RVCEs were generated. Results are shown in \cref{tab:diversity}. Naturally, decreasing the area hyperparameter limits the extent of possible changes, leading to a decrease in diversity. Picking $a=0.3$ results in diversity comparable to values reported by previous works, \eg, \citet{jeanneret2023adversarial}.

\begin{table}[H]
    \centering
    \scriptsize
    \begin{tabular}{l|c|c|c}
        \toprule
         & \textbf{Zebra -- Sorrel} & \textbf{Cheetah -- Cougar} & \textbf{Egyptian Cat – Persian Cat} \\
        \midrule
         \method{}$^A$ & $0.067$ & $0.060$ & $0.065$ \\
         \method{}$^B$ & $0.092$ & $0.096$ & $0.095$ \\
         \method{}$^C$ & $0.129$ & $0.140$ & $0.137$ \\
        \bottomrule
    \end{tabular}
    \caption{Diversity evaluation using $256$ images for each task from our experimental protocol.}
    \label{tab:diversity}
\end{table}

\subsection{Computational efficiency assessment}

What connects previous SOTA SGM-based methods with our work is the use of large U-Net \citep{ronneberger2015u} checkpoints for the denoising network with the number of hyperparameters far exceeding (\eg, $10\times$) the size of the utilized classifier, effectively dominating the computational burden. Hence, to ensure fair comparison, \cref{table:nfe_grid} shows the number of Neural Function Evaluations (NFEs) used by each method to produce a single explanation, divided into a. model (\emph{U-Net}, \emph{classifier} and \emph{other}) b. and \emph{forward} / \emph{backward} passes, where the backward pass is around $2\times$ more computationally demanding than the forward pass. Importantly, this comparison eliminates the differences stemming from the utilized hardware and optimality of the implementation, and is fair as each method consumes virtually the same amount of GPU memory. One exception to the use of the standard U-Net model is the LDCE method of \cite{farid2023latent}, which applies it in the latent space of an autoencoder. However, as each latent U-Net step also requires decoding the image with the decoder, the computational demand stays similar to the standard approach.

As indicated by \cref{table:nfe_grid}, \method{} is the most efficient approach, both in terms of balancing the use of the U-Net and the classifier, and the number of forward/backward passes. Importantly, the \emph{other} category shows non-zero numbers only for the DVCE method, which additionaly uses the gradients of a robust classifier in the generation process. The high number of forward/backward passes through the classifiers in DVCE stems from applying them to a set of $16$ augmented versions of $\mathbf{x}_t$ at each timestep.

\begin{table}[ht]
\centering
    \scriptsize
    \begin{tabular}{c|cc|cc|cc}
    \toprule
    \multicolumn{7}{c}{\textbf{NFE}} \\
    \midrule
    \multirow{2}{*}{Inpainting method} & \multicolumn{2}{c}{U-Net} & \multicolumn{2}{c}{Classifier} & \multicolumn{2}{c}{Other} \\
    \cmidrule(lr){2-7}
    & forward & backward & forward & backward & forward & backward   \\
    \midrule
    \method{} & $100$ & $100$ & $100$ & $100$ & $0$ & $0$ \\
    LDCE      & $191$ & $191$ & $191$ & $191$ & $0$ & $0$ \\
    DVCE      & $200$ & $200$ & $1600$ & $1600$ & $1600$ & $1600$ \\
    ACE       & $520$ & $500$ & $25$ & $25$ & $0$ & $0$ \\
    DDRM      & $200$ & $200$ & $200$ & $200$ & $0$ & $0$ \\
    MCG       & $1000$ & $1000$ & $1000$ & $1000$ & $0$ & $0$ \\
    RePaint   & $2410$ & $2410$ & $2410$ & $2410$ & $0$ & $0$ \\
    \bottomrule
    \end{tabular}
    \caption{Number of NFEs for each respective method with details about the model type and forward/backward passes.}
    \label{table:nfe_grid}
\end{table}

\subsection{Additional quantitative results}

For most visually appealing results, we found $a \approx 0.1-0.15$, $c\approx 8-16$, $\tau \approx 0.3-0.6$ and $s \approx 2-3$ to perform the best. These hyperparameters were used to create RVCEs for \cref{fig:4_main_automated}. To assess that the performance of these configurations does not deviate from best configurations of \cref{tab:main_results}, we followed the same evaluation protocol on the most challenging Zebra--Sorrel task and include the results in \cref{tab:figure-metrics}. Crucially, the performance stays virtually the same when comparing with \cref{tab:main_results}.

\begin{table}[H]
\centering
    \scriptsize
    \begin{tabular}{*{4}{c|}*{5}{c}}
    \toprule
     Area $a$ & Cell size $c$ & Guidance scale $s$ & Trajectory truncation $\tau$ & FID & sFID & S$^3$ & COUT & FR \\ 
    \midrule
    \multirow{9}{*}{$0.1$} & \multirow{9}{*}{$16$} & \multirow{3}{*}{$3.0$} & $0.2$ & $10.1$ & $18.4$ & $0.92$ & $0.79$ & $95.8$ \\
                     &                     &                      & $0.3$ & $10.7$ & $19.0$ & $0.91$ & $0.76$ & $95.0$ \\
                     &                     &                      & $0.4$ & $10.8$ & $18.9$ & $0.90$ & $0.74$ & $94.3$ \\
\cmidrule{3-9}
                     &                     & \multirow{3}{*}{$3.5$} & $0.2$ & $11.0$ & $19.2$ & $0.91$ & $0.81$ & $97.0$ \\
                     &                     &                      & $0.4$ & $11.4$ & $19.4$ & $0.89$ & $0.77$ & $96.0$ \\
                     &                     &                      & $0.3$ & $11.6$ & $19.7$ & $0.90$ & $0.79$ & $96.2$ \\
\cmidrule{3-9}
                     &                     & \multirow{3}{*}{$4.0$} & $0.2$ & $11.7$ & $19.8$ & $0.91$ & $0.83$ & $97.2$ \\
                     &                     &                      & $0.4$ & $12.3$ & $20.2$ & $0.88$ & $0.79$ & $96.7$ \\
                     &                     &                      & $0.3$ & $12.4$ & $20.4$ & $0.89$ & $0.80$ & $96.2$ \\
    \midrule
\multirow{9}{*}{$0.15$} & \multirow{9}{*}{$16$} & \multirow{3}{*}{$2.0$} & $0.4$ & $11.2$ & $19.2$ & $0.89$ & $0.77$ & $97.7$ \\
                      &                     &                    & $0.5$ & $12.1$ & $20.0$ & $0.88$ & $0.74$ & $96.5$ \\
                      &                     &                    & $0.6$ & $12.7$ & $20.4$ & $0.86$ & $0.70$ & $94.8$ \\
\cmidrule{3-9}
                      &                     & \multirow{3}{*}{$3.0$} & $0.4$ & $13.5$ & $21.3$ & $0.87$ & $0.82$ & $99.5$ \\
                      &                     &                    & $0.5$ & $13.9$ & $21.7$ & $0.86$ & $0.80$ & $99.2$ \\
                      &                     &                    & $0.6$ & $14.2$ & $21.7$ & $0.85$ & $0.78$ & $98.6$ \\
\cmidrule{3-9}
                      &                     & \multirow{3}{*}{$4.0$} & $0.4$ & $15.3$ & $22.9$ & $0.86$ & $0.84$ & $99.6$ \\
                      &                     &                    & $0.5$ & $15.5$ & $23.1$ & $0.85$ & $0.82$ & $99.7$ \\
                      &                     &                    & $0.6$ & $15.8$ & $23.5$ & $0.83$ & $0.81$ & $99.4$ \\
    \bottomrule
    \end{tabular}
    \caption{Quantitative results for hyperparameters that provide the most visually appealing results.}
    \label{tab:figure-metrics}
\end{table}

\section{Qualitative examples}\label{app:qualitative}

\begin{figure}
    \centering
    \includegraphics[width=1\linewidth]{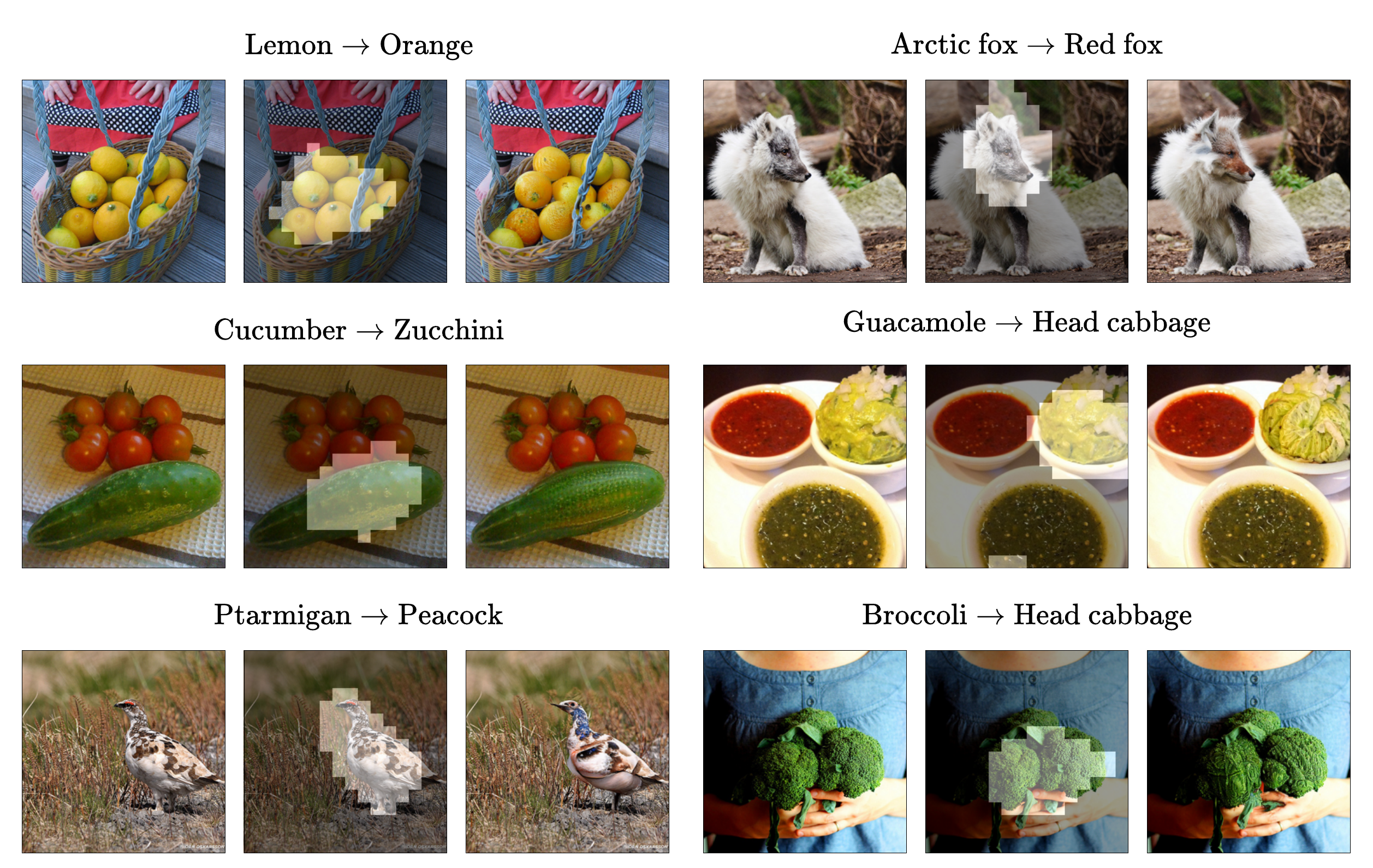}
    \caption{Extended qualitative evaluation of automated region extraction for VGG16 \citep{simonyan2014very} classifier. For each task, factual image is shown on the left with the used region in the middle and the generated RVCE on the right.}
    \label{fig:automated_vgg16}
\end{figure}

\begin{figure}
    \centering
    \includegraphics[width=1\linewidth]{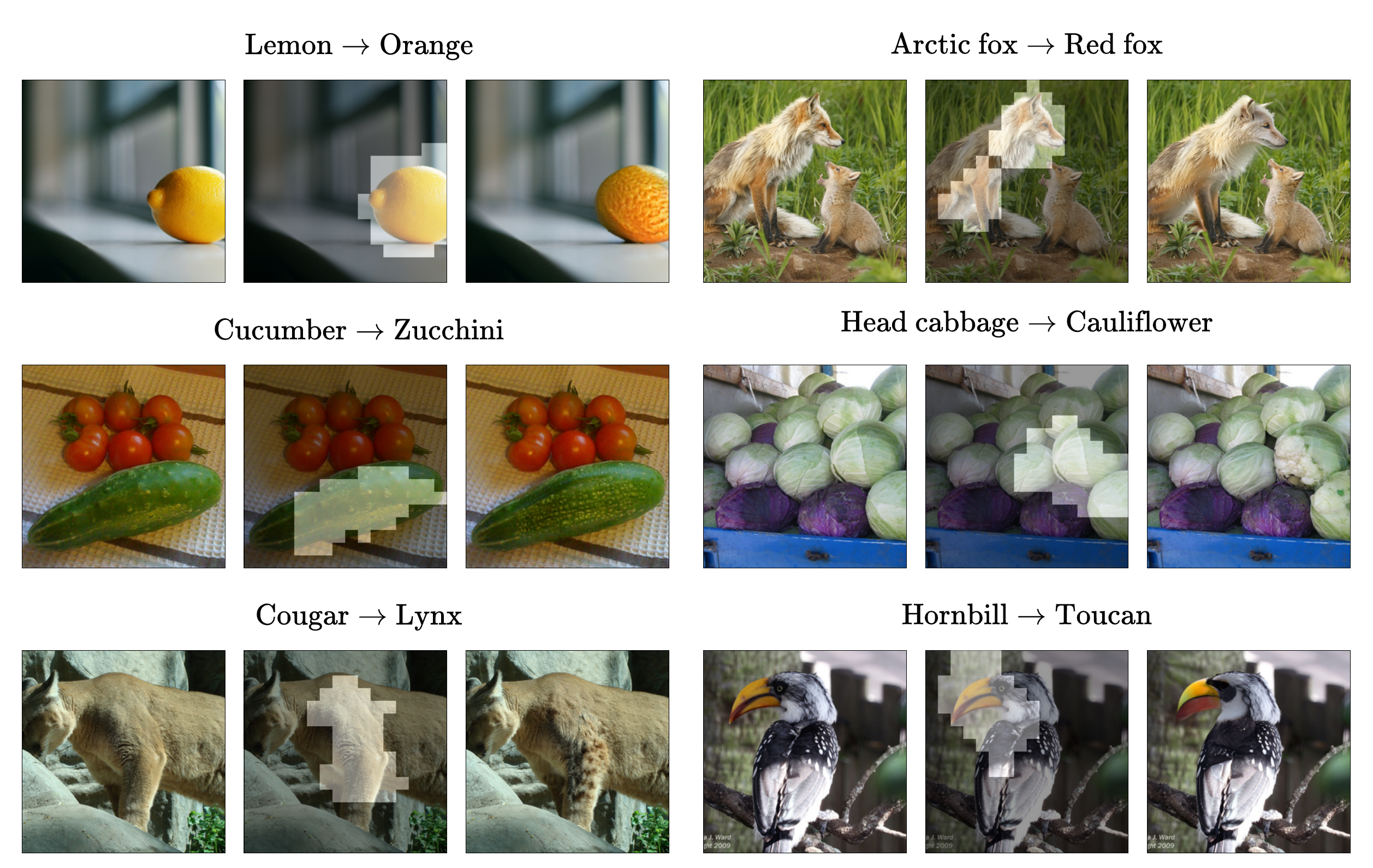}
    \caption{Extended qualitative evaluation of automated region extraction for VGG16BN \citep{simonyan2014very} classifier. For each task, factual image is shown on the left with the used region in the middle and the generated RVCE on the right.}
    \label{fig:automated_vgg16bn}
\end{figure}

\begin{figure}
    \centering
    \includegraphics[width=1\linewidth]{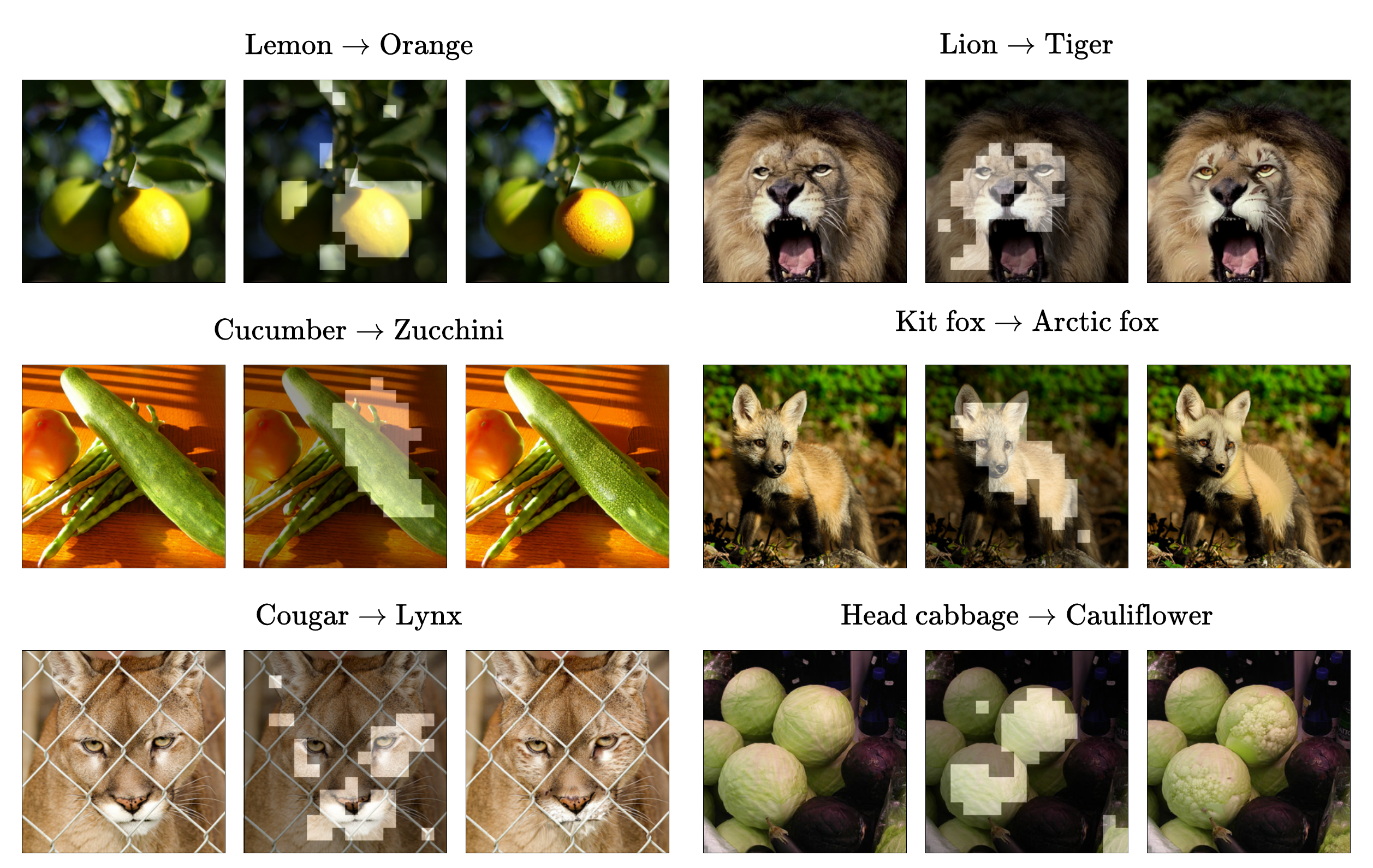}
    \caption{Extended qualitative evaluation of automated region extraction for ConvNeXt Base \citep{liu2022convnet} classifier. For each task, factual image is shown on the left with the used region in the middle and the generated RVCE on the right.}
    \label{fig:automated_convnextbase}
\end{figure}

\begin{figure}
    \centering
    \includegraphics[width=1\linewidth]{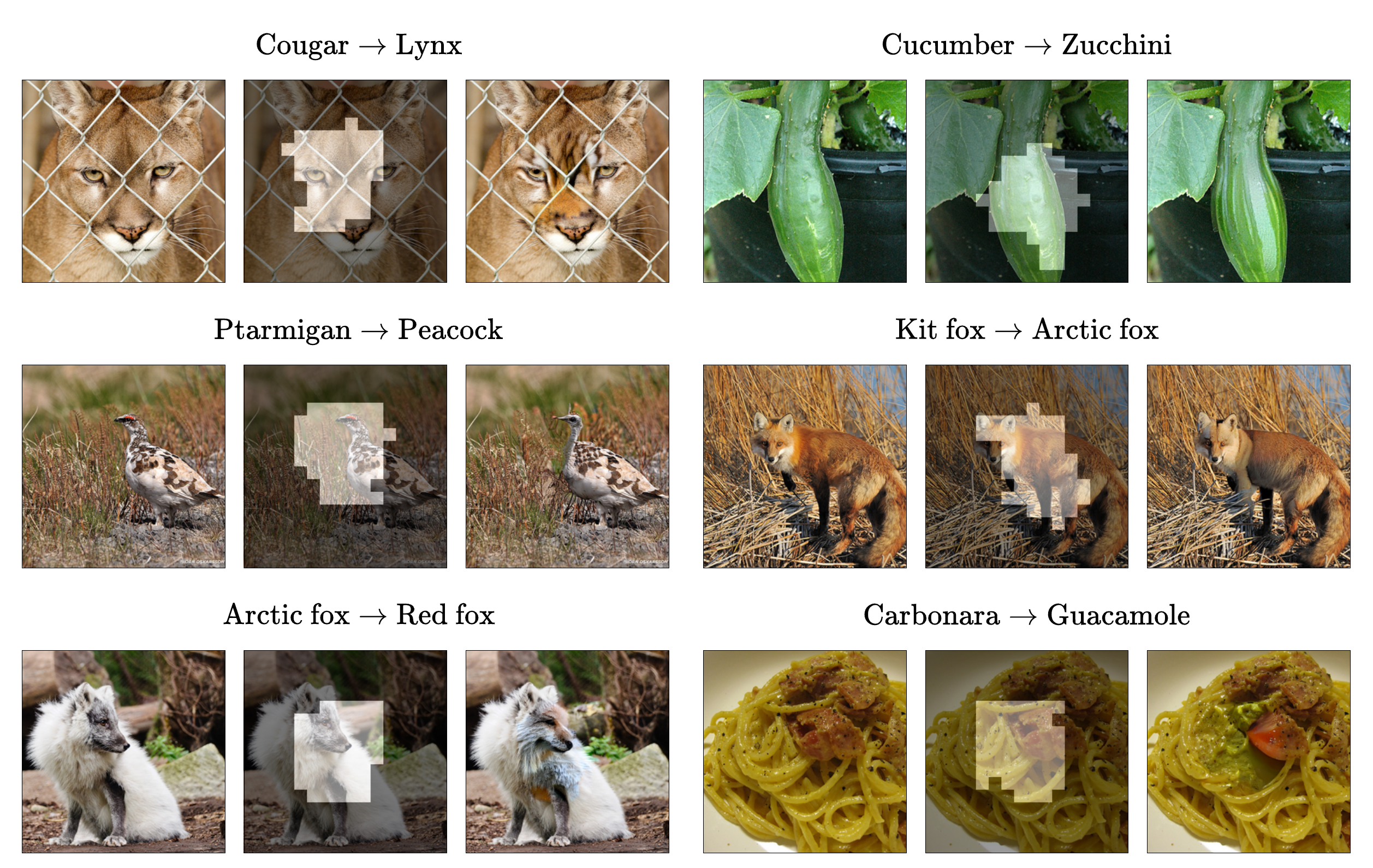}
    \caption{Extended qualitative evaluation of automated region extraction for ViTB16 \citep{dosovitskiy2020image} classifier. For each task, factual image is shown on the left with the used region in the middle and the generated RVCE on the right.}
    \label{fig:automated_vit}
\end{figure}

\begin{figure}
    \centering
    \includegraphics[width=1\linewidth]{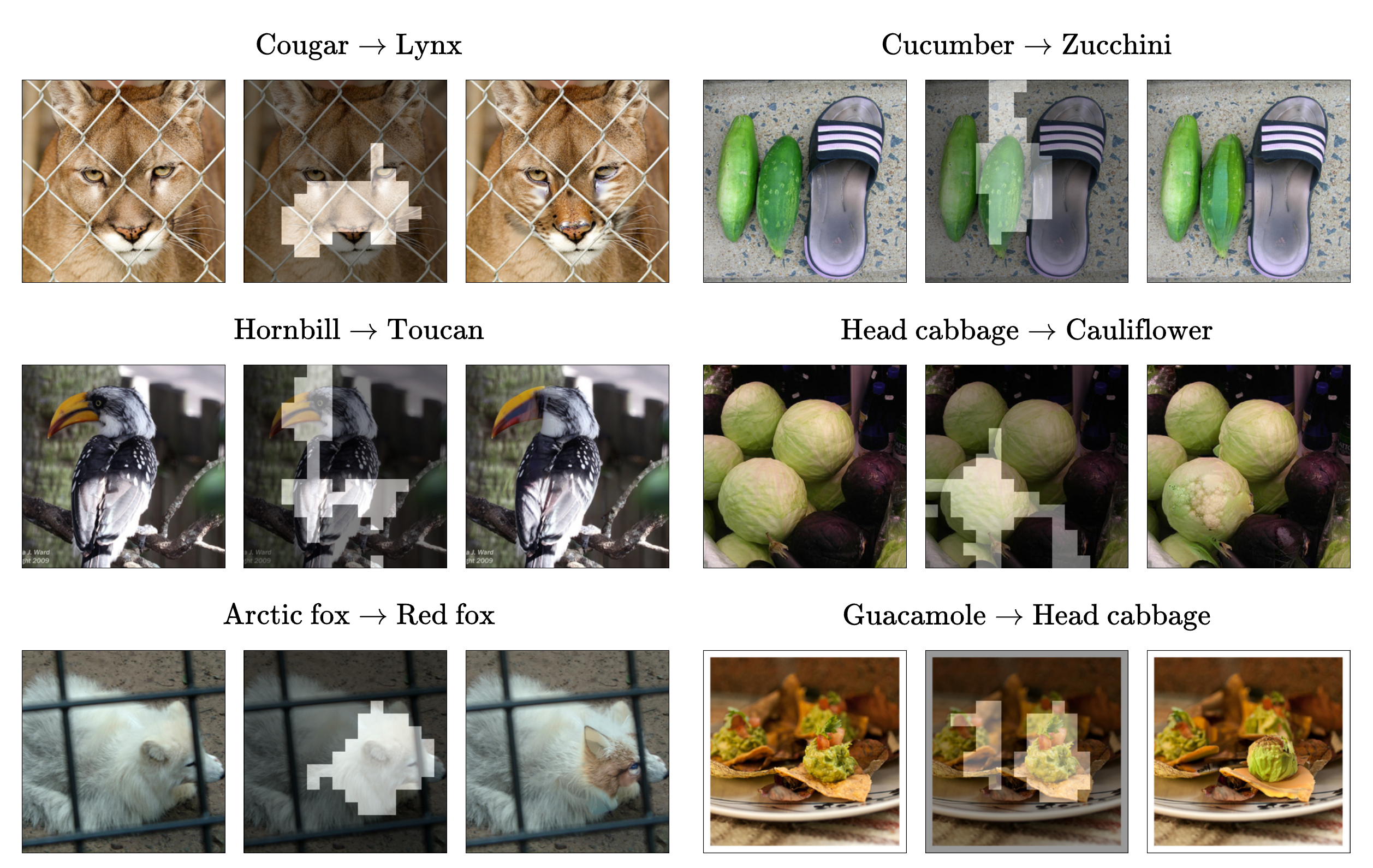}
    \caption{Extended qualitative evaluation of automated region extraction for SwinB \citep{liu2021swin} classifier. For each task, factual image is shown on the left with the used region in the middle and the generated RVCE on the right.}
    \label{fig:automated_swinb}
\end{figure}

\begin{figure}
    \centering
    \includegraphics[width=1\linewidth]{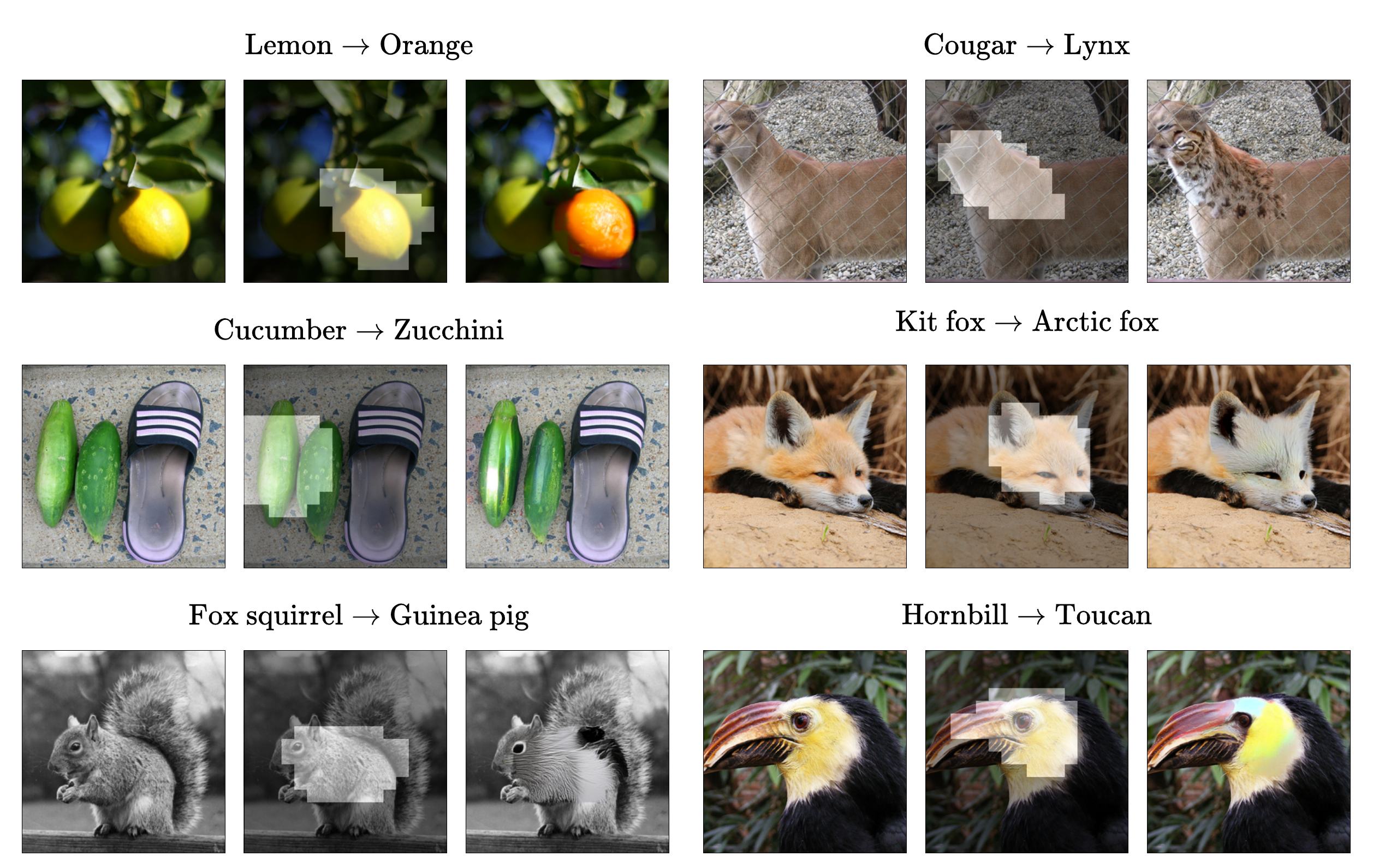}
    \caption{Extended qualitative evaluation of automated region extraction for Madry ResNet50 \citep{robustness} $l_2$-norm robust classifier. For each task, factual image is shown on the left with the used region in the middle and the generated RVCE on the right.}
    \label{fig:automated_madryresnet50}
\end{figure}

\begin{figure}
    \centering
    \includegraphics[width=1\linewidth]{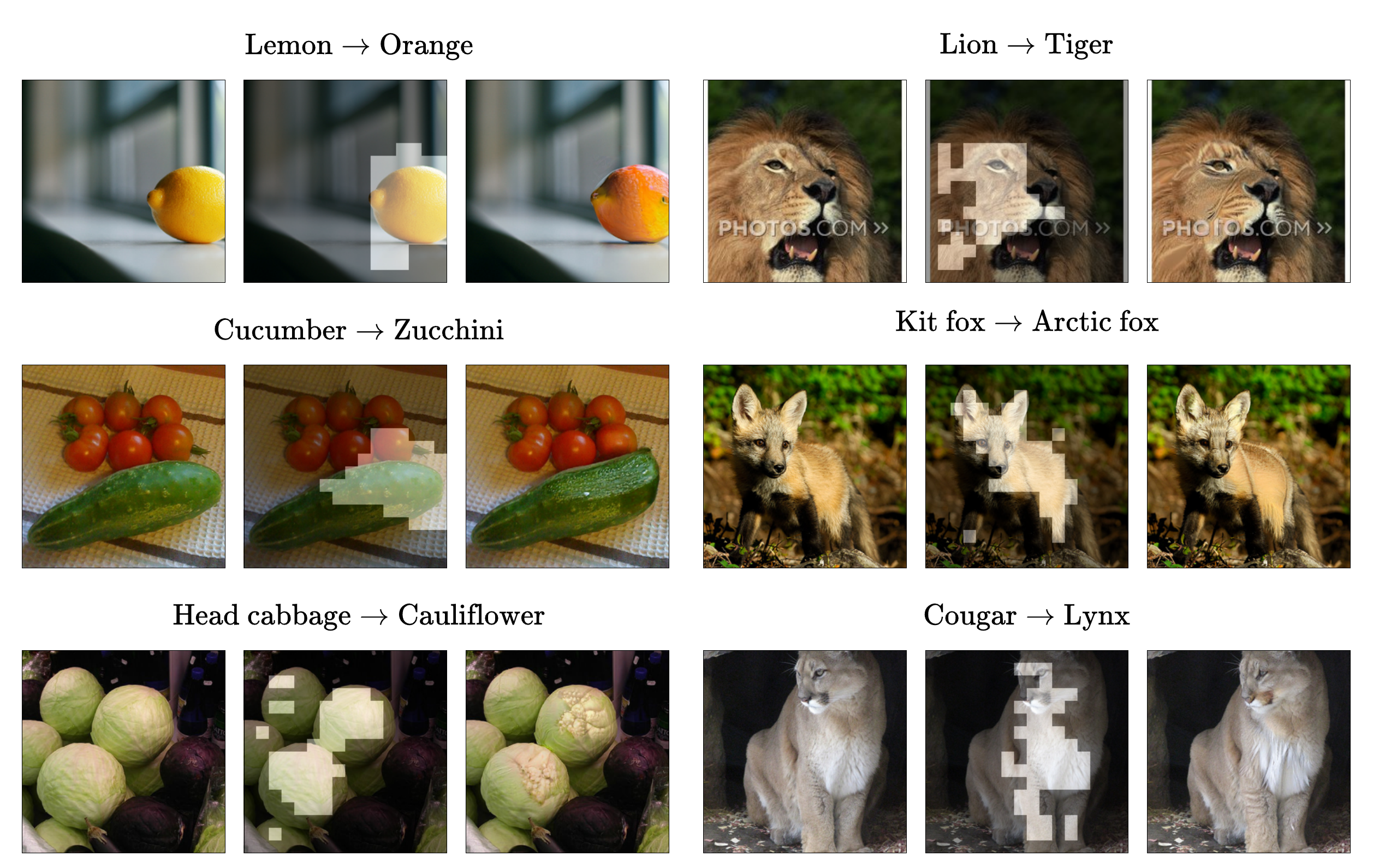}
    \caption{Extended qualitative evaluation of automated region extraction for Tian DeiT \citep{tian2022deeper} corruption robust classifier. For each task, factual image is shown on the left with the used region in the middle and the generated RVCE(s) on the right.}
    \label{fig:automated_robustdeit}
\end{figure}

\begin{figure}
    \centering
    \includegraphics[width=1\linewidth]{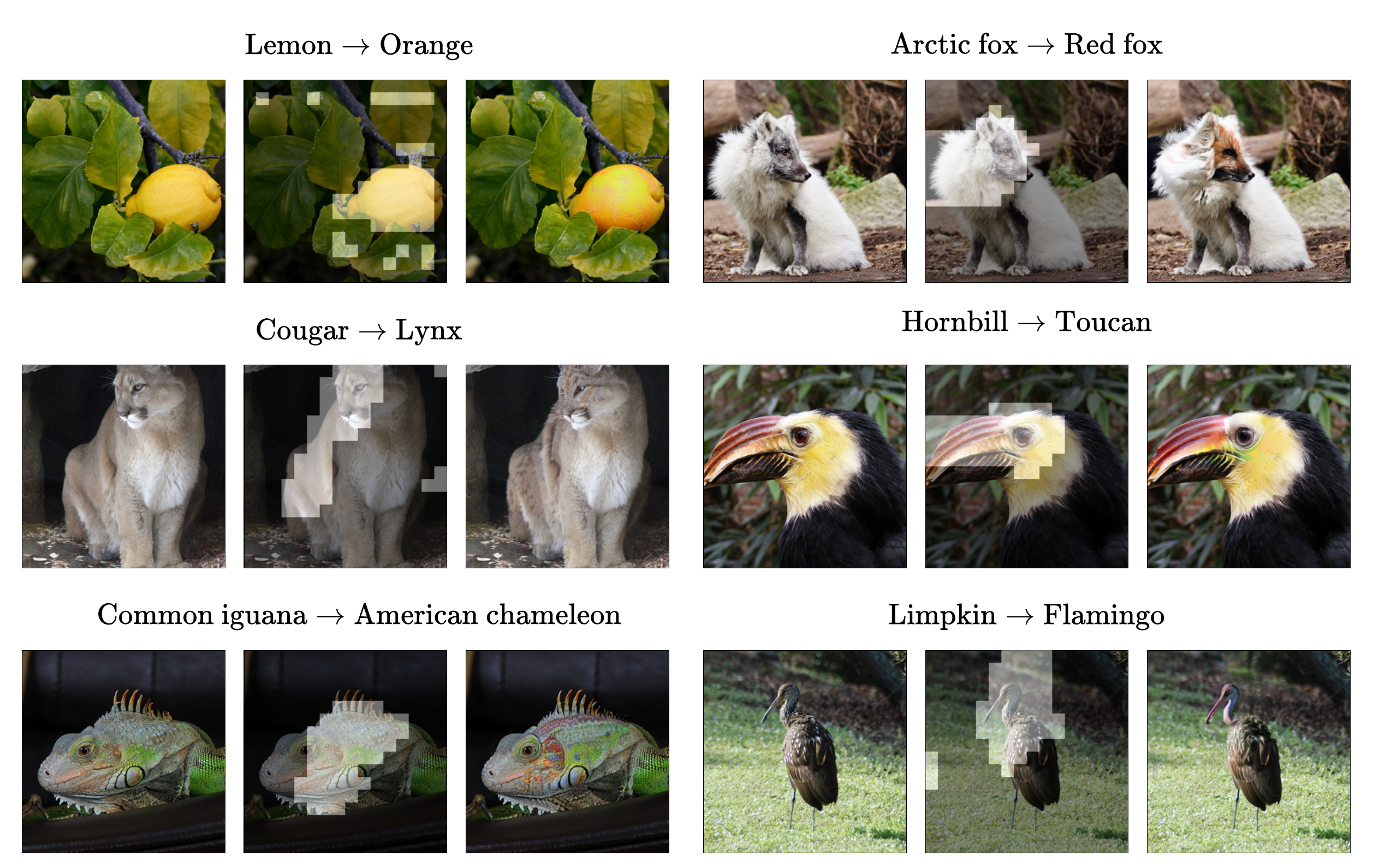}
    \caption{Extended qualitative evaluation of automated region extraction for CLIP ViT-B/32 \citep{radford2021learning} zero-shot classifier. For each task, factual image is shown on the left with the used region in the middle and the generated RVCE on the right.}
    \label{fig:automated_clipzeroshot}
\end{figure}

\begin{figure}
    \centering
    \includegraphics[width=1.0\linewidth]{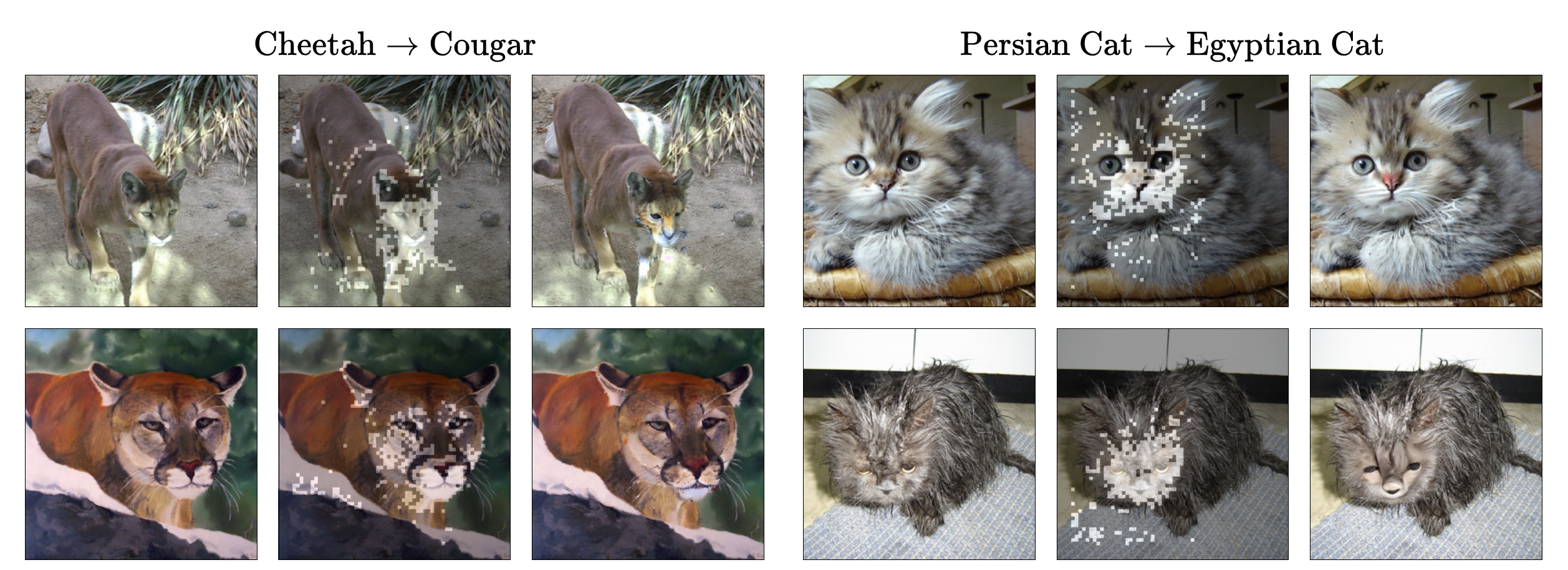}
    \caption{Extended qualitative evaluation of automated region extraction with $c=4$, $a=0.1$ for the ResNet50 classifier. For each task, factual image is shown on the left with the used region in the middle and the generated RVCE on the right.}
    \label{fig:automated_cell_size_4_area_0_1}
\end{figure}

\begin{figure}
    \centering
    \includegraphics[width=1.0\linewidth]{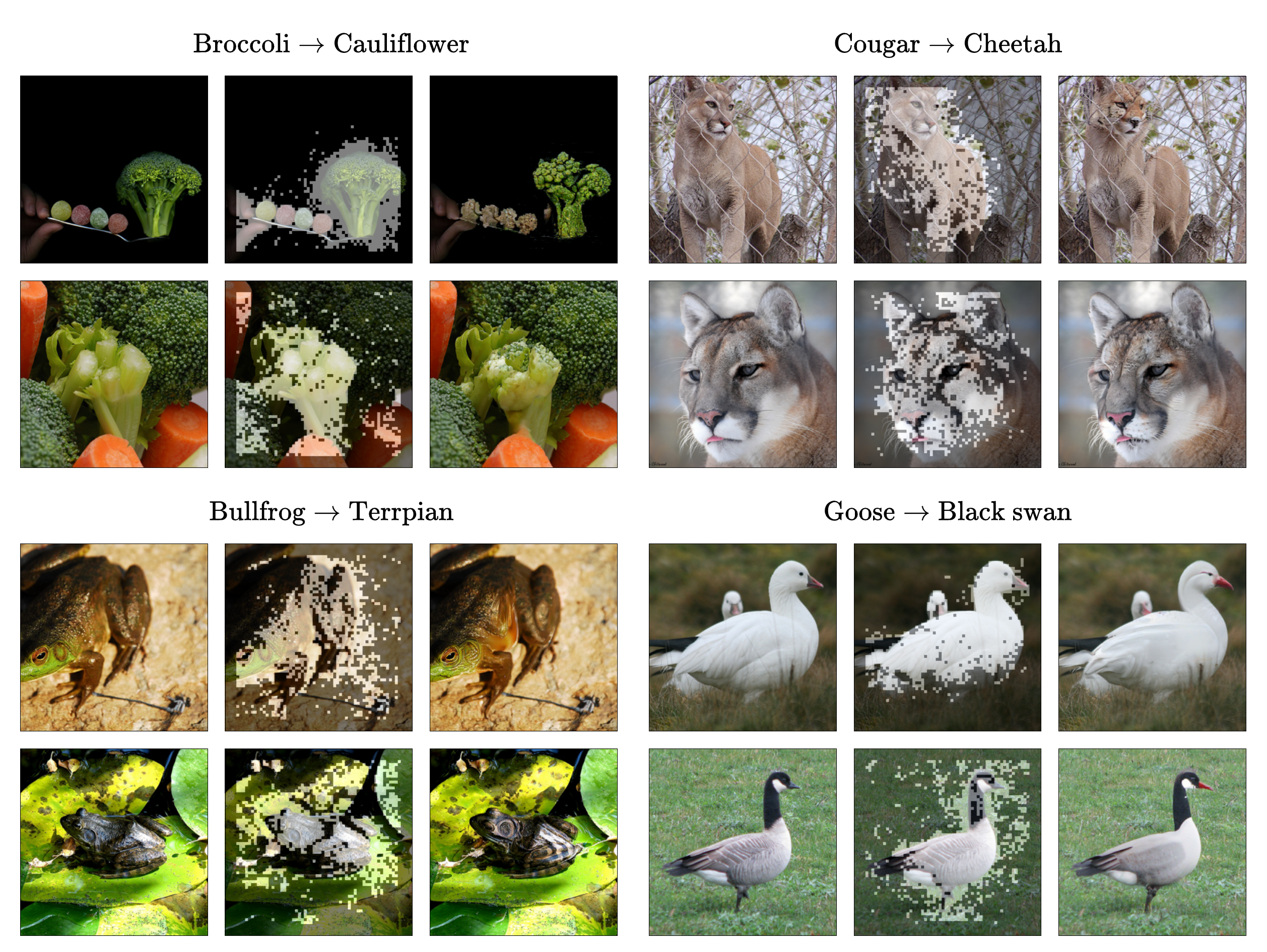}
    \caption{Extended qualitative evaluation of automated region extraction with $c=4$, $a=0.3$ for the ResNet50 classifier. For each task, factual image is shown on the left with the used region in the middle and the generated RVCE on the right.}
    \label{fig:automated_cell_size_4_area_0_3}
\end{figure}

\begin{figure}
    \centering
    \includegraphics[width=1.0\linewidth]{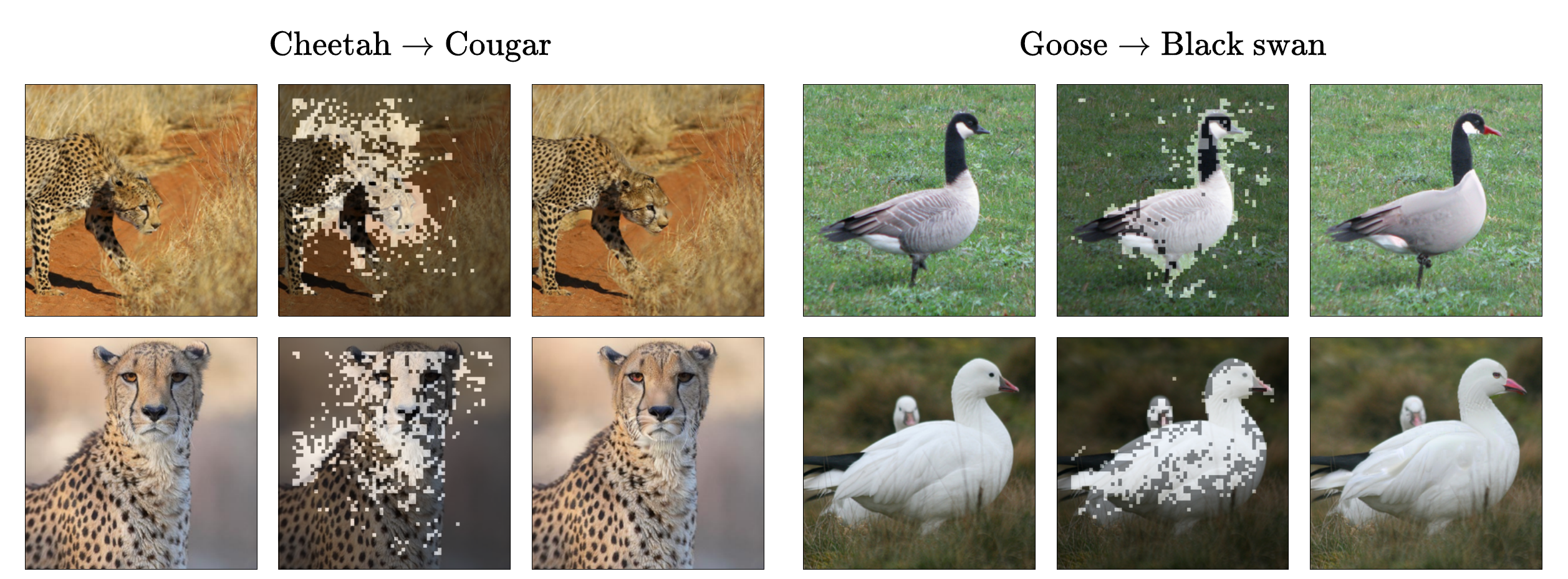}
    \caption{Extended qualitative evaluation of automated region extraction with $c=4$, $a=0.2$ for the ResNet50 classifier. For each task, factual image is shown on the left with the used region in the middle and the generated RVCE on the right.}
    \label{fig:automated_cell_size_4_area_0_2}
\end{figure}

\begin{figure}
    \centering
    \includegraphics[width=1.0\linewidth]{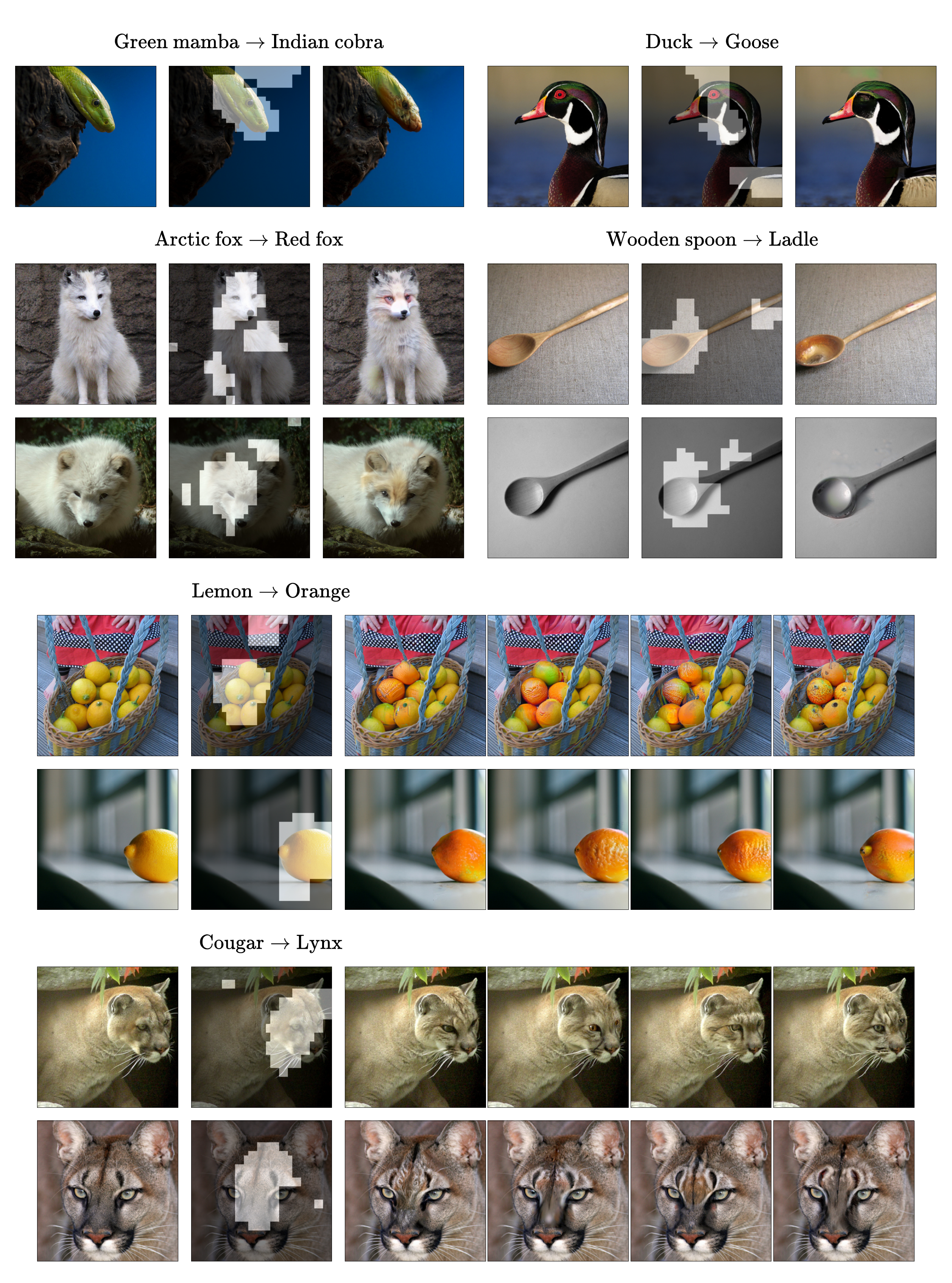}
    \caption{Extended qualitative evaluation of automated region extraction with $c=8$, $a=0.2$ for the ResNet50 classifier. For each task, factual image is shown on the left with the used region in the middle and the generated RVCE(s) on the right.}
    \label{fig:automated_cell_size_8_area_0_2}
\end{figure}

\begin{figure}
    \centering
    \includegraphics[width=1.0\linewidth]{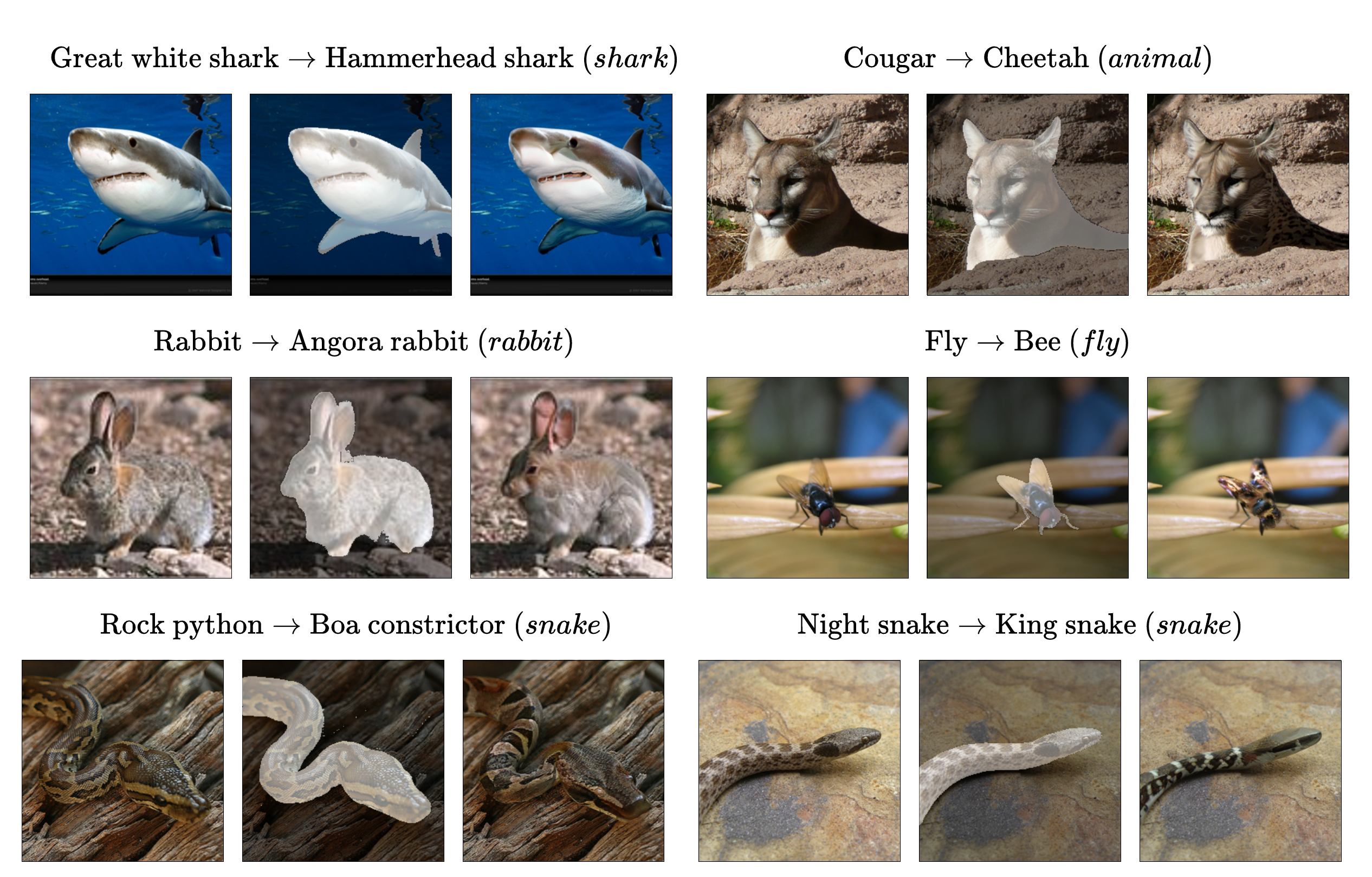}
    \caption{Extended qualitative evaluation of exact regions obtained with LangSAM for the ResNet50 classifier. For each task, factual image is shown on the left with the used region in the middle and the generated RVCE(s) on the right.}
    \label{fig:langsam_appendix}
\end{figure}

\begin{figure}
    \centering
    \includegraphics[width=1.0\linewidth]{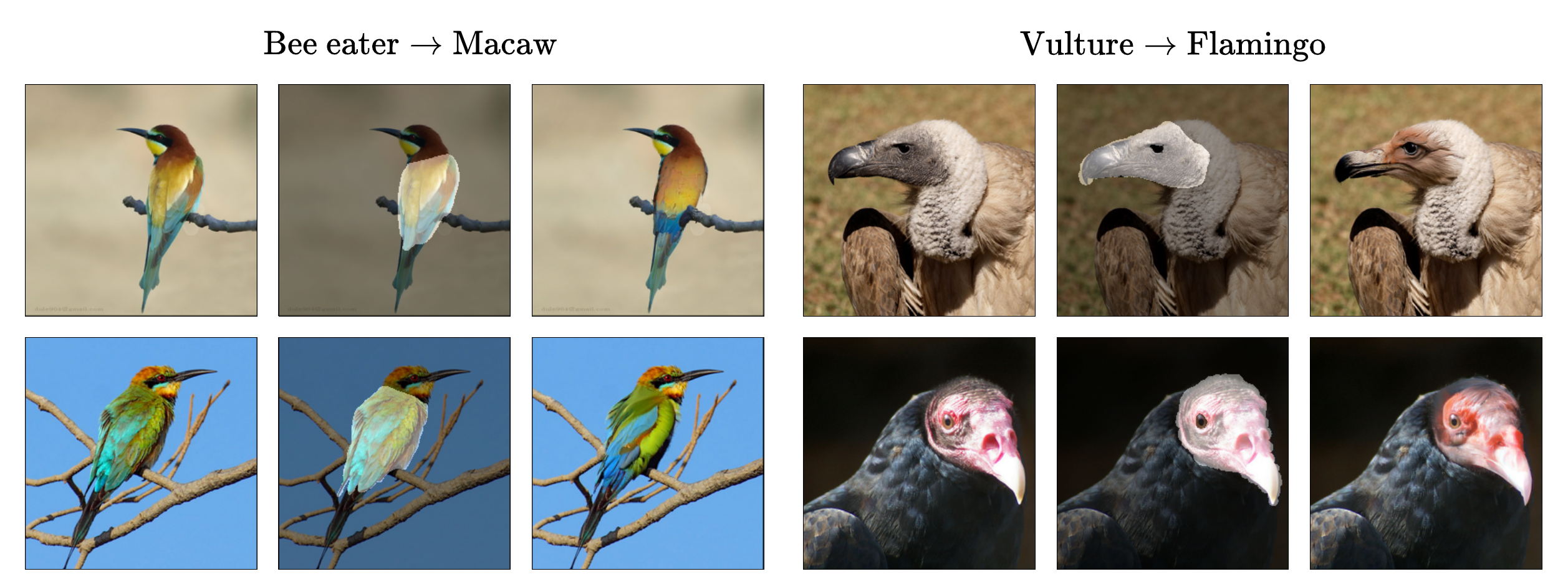}
    \caption{Extended qualitative evaluation of user-defined regions for the ResNet50 classifier. For each task, factual image is shown on the left with the used region in the middle and the generated RVCE(s) on the right.}
    \label{fig:user_defined_appendix}
\end{figure}

We provide additional qualitative examples for different scenarios. Results for other classifiers, obtained with the automated region extraction, are depicted in \cref{fig:automated_vgg16} (VGG16, \citet{simonyan2014very}), \cref{fig:automated_vgg16bn} (VGG16 with Batch Normalization (BN), \citet{simonyan2014very}), \cref{fig:automated_convnextbase} (ConvNeXt Base, \citet{liu2022convnet}), \cref{fig:automated_vit} (ViT-B/16, \citet{dosovitskiy2020image}), \cref{fig:automated_swinb} (SwinB, \citet{liu2021swin}) , \cref{fig:automated_madryresnet50} (robust Madry ResNet50, \citet{robustness}), \cref{fig:automated_robustdeit} (robust Tian DeiT, \citet{tian2022deeper}), \cref{fig:automated_clipzeroshot} (zero-shot CLIP classifier, \citet{radford2021learning}). We used $a=0.2,\tau=0.7,c=16,s=1.5$ universally across all additional classifiers, showcasing the versatility of \method{}. Additional examples for ResNet50 are shown in \cref{fig:automated_cell_size_4_area_0_2,fig:automated_cell_size_8_area_0_2,fig:automated_cell_size_4_area_0_3,fig:automated_cell_size_4_area_0_1} (automated region extraction with different hyperparameters), \cref{fig:langsam_appendix} (exact regions from LangSAM) and \cref{fig:user_defined_appendix} (user-defined regions).

\newpage

\end{document}